\documentclass{article}
\usepackage{iclr2027,times}

\usepackage{amsmath,amsfonts,bm}

\def\eqref#1{equation~\ref{#1}}

\def\1{\bm{1}}

\DeclareMathAlphabet{\mathsfit}{\encodingdefault}{\sfdefault}{m}{sl}
\SetMathAlphabet{\mathsfit}{bold}{\encodingdefault}{\sfdefault}{bx}{n}

\usepackage{graphicx}
\usepackage{microtype}
\usepackage{booktabs}
\usepackage[table]{xcolor}
\usepackage{arydshln}
\usepackage{amsmath}
\usepackage{amssymb}
\usepackage{mathtools}
\usepackage{subcaption}
\usepackage{placeins}
\usepackage[most]{tcolorbox}
\tcbuselibrary{breakable,listings}

\usepackage{hyperref}
\usepackage{url}
\usepackage[capitalize,noabbrev]{cleveref}

\renewcommand{\arraystretch}{1.45}

\title{Imagine to Act: High-Fidelity Data Synthesis via Image Editing
World Model for Scalable GUI Agent Training}

\author{
Yongxin Ning\textsuperscript{1}, Runliang Niu\textsuperscript{1}, Qianli Xing\textsuperscript{2},
Zhiyi Duan\textsuperscript{2}, Qingzu He\textsuperscript{3}, Pan Wang\textsuperscript{3}, \\
\bfseries Qi Wang\textsuperscript{1}\\
\textsuperscript{1}School of Artificial Intelligence, Jilin University\\
\textsuperscript{2}College of Computer Science, Jilin University\\
\textsuperscript{3}OPPO\\
\texttt{ningyx25@mails.jlu.edu.cn}, \texttt{qiwang@jlu.edu.cn}
}

\makeatletter
\def\@maketitle{\vbox{\hsize\textwidth
{\LARGE \@title\par}%
    \lhead{Preprint}%
    \def\And{\end{tabular}\hfil\linebreak[0]\hfil
            \begin{tabular}[t]{l}\bf\rule{\z@}{24pt}\ignorespaces}%
  \def\AND{\end{tabular}\hfil\linebreak[4]\hfil
            \begin{tabular}[t]{l}\bf\rule{\z@}{24pt}\ignorespaces}%
    \begin{tabular}[t]{l}\bf\rule{\z@}{24pt}\@author\end{tabular}%
\vskip 0.3in minus 0.1in}}
\makeatother

\iclrfinalcopy
\begin{document}

\maketitle

\begin{abstract}\label{sec:abstract}
Graphical User Interface (GUI) agents have emerged as a promising paradigm for automating complex digital workflows across diverse applications. However, training highly capable and generalizable agents fundamentally relies on massive, high-fidelity visual-action trajectories, which are notoriously difficult to acquire. While human demonstrations are unscalable, existing GUI world models rely on text descriptions or HTML rendering, discarding crucial pixel-level visual details like icons and layout styles. To address this issue, we introduce Infinite-Dreamer, a simulation-free data synthesis method powered by a pixel-level Image Editing World Model. By conceptualizing GUI transitions as image editing tasks, we leverage Vision-Language Models (VLMs) to describe action-induced UI changes as structured delta-text. We then fine-tune an image editing backbone to controllably synthesize realistic screenshot transitions. We utilize this model to generate both single-frame visual robustness data and multi-step imaginary trajectories. To validate the effectiveness of our approach, we fine-tune the Qwen3-VL baseline solely on the synthesized data to obtain Infinite-Actor, and evaluate it on AndroidWorld, MobileWorld, and AndroidControl-Curated benchmarks. Infinite-Actor consistently outperforms the Qwen3-VL baselines across scales: Infinite-Actor-8B improves AndroidWorld Pass@1 by +4.45 and nearly doubles the MobileWorld Pass@3 success rate, while Infinite-Actor-2B improves Pass@1 by +9.05. Code is available at \url{https://github.com/swaydy-n/Infinite-Dreamer}.
\end{abstract}

\section{Introduction}\label{sec:introduction}

Graphical User Interface (GUI) agents have rapidly emerged as a pivotal frontier in artificial intelligence, promising to seamlessly automate increasingly complex digital workflows across diverse operating systems and web applications. Unlike text-based language agents, GUI agents must perceive rich visual screens, effectively reason over layout structures, and accurately execute precise actions just like human users. To achieve robust generalizability across infinitely varied real-world digital environments, these agents must master the intricate mapping between natural language instructions, dynamic screen layouts, and sequential action executions. Ultimately, the development of such highly capable and adaptable GUI agents is fundamentally contingent upon the availability of massive, diverse, and high-fidelity annotated visual-action training data.

Despite its critical importance, acquiring high-quality trajectory data annotated with explicit action intents remains exceptionally difficult. Existing efforts to obtain such data generally suffer from two major limitations: (i) \textit{Human-driven demonstration} is not only financially prohibitive and labor-intensive, but it is also severely bottlenecked by environmental constraints, such as the technical friction of configuring heterogeneous device platforms and the inherent risks of exposing sensitive user privacy during the recording process; (ii) \textit{Automated environment exploration} typically relies on heuristic search or random-walk strategies. Lacking a long-term semantic objective, these automated agents easily become trapped in trivial UI loops, frequently trigger unrecoverable error dead-ends, or generate disjointed action sequences. Consequently, such exploration-based approaches heavily bias the collected data toward short, repetitive interactions, making them incapable of scaling to complex, multi-step workflows and utterly failing to maintain goal coherence over long execution sequences. Therefore, how to scalably synthesize GUI interaction trajectories while improving data efficiency has emerged as a core challenge in training advanced GUI agents.

\begin{figure}[tb]
\centering
\includegraphics[width=\columnwidth]{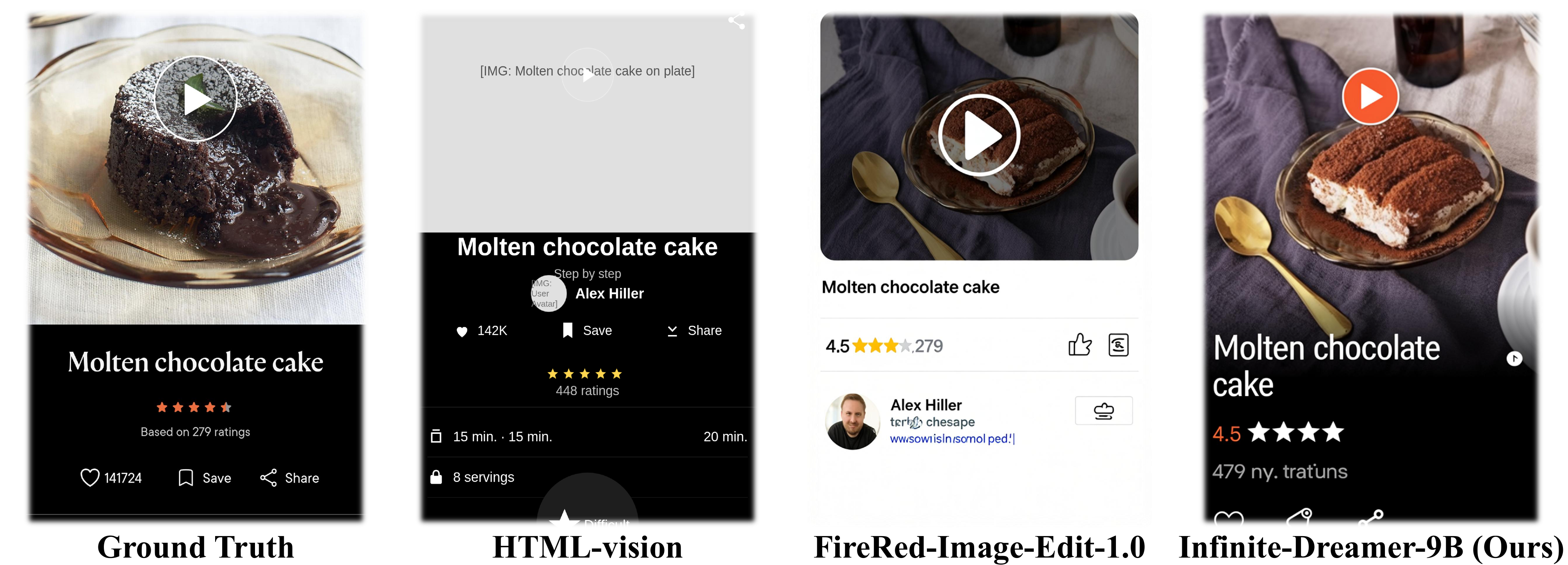}
\caption{\textbf{Motivation.} Post-action screenshots predicted by different world models: (1) ground truth; (2) HTML-vision rendering, lacking visual detail; (3) the pretrained FireRed-Image-Edit-1.0, failing to capture GUI-specific transitions; and (4) our Infinite-Dreamer-9B, producing high-fidelity, pixel-level screenshots faithful to the ground truth.}
\label{fig:motivation}
\end{figure}

To address this challenge, the paradigm of world models offers a powerful solution for data synthesis. Fundamentally, world models are specifically designed to internalize the underlying transition dynamics of an environment, predicting subsequent states by learning the causal laws that govern how the world responds to specific actions. In the digital ecosystem, a robust GUI world model acts as an unconstrained virtual simulator, emulating how software applications evolve after a click or a scroll. While recent pioneering GUI world models, such as Qwen-AgentWorld and Code2World, attempt to simulate environments via text descriptions or HTML rendering, they inherently struggle to capture diverse layouts and fine-grained visual features, such as custom icons, banners, and dynamic animations (Figure \ref{fig:motivation}). In contrast, pixel-level screenshots represent the natural, raw state space of GUIs, directly reflecting the visual reality experienced by human users. A pixel-level world model can implicitly learn the visual regularities of UI evolution, such as a loading spinner disappearing or a drop-down menu unfolding, thereby preserving both rich graphical details and complex software execution logic. Therefore, leveraging image editing models directly as GUI world models offers a more natural, direct, and high-fidelity paradigm for realistic data synthesis.

To this end, we introduce Infinite-Dreamer, a high-fidelity, simulation-free data synthesis pipeline that generates pixel-level screenshots and continuous action trajectories without relying on complex and restrictive simulation environments. First, we construct a specialized image editing dataset built upon AndroidControl. By employing a Vision-Language Model (VLM) to describe the transitions of UI elements before and after an action, we generate fine-grained delta-text that precisely captures added or removed interface elements. Second, we apply fine-tuning to a generative image editing backbone. The resulting model serves as a pixel-level GUI world model, capable of editing a pre-action screenshot into a post-action screenshot conditioned on the delta-text. Finally, to demonstrate how our synthesized data enhances downstream agent performance, we generate two distinct categories of samples: (i) visual robustness data, where we augment single screenshots with diverse layouts, colors, and styles to boost the agent's adaptability, and (ii) multi-step imaginary trajectories, where the world model continuously edits an initial screenshot to "imagine" multi-step task executions, expanding state coverage beyond the limits of real-world datasets. To summarize, our major contributions are:

\begin{itemize}
    \item \textbf{A Controllable, Pixel-Level GUI World Model.} We repurpose an image editing model into a screenshot-to-screenshot GUI world model conditioned on structured delta-text. This approach effectively enables the controllable generation of diverse, high-fidelity GUI screens without relying on complex costly simulators.
    \item \textbf{Enhanced Data Efficiency and Generalization.} We demonstrate that our synthesis model significantly boosts the utility of existing trajectories. By blending synthetic and real-world data, the trained agent generalizes far better to unseen UI layouts and novel interaction patterns, outperforming agents trained purely on real demonstrations.
    \item \textbf{Substantial Downstream Performance Gains.} We train a vision-language agent, called Infinite-Actor, using our newly synthesized data. Evaluation on several challenging online benchmarks reveals substantial dramatic performance gains; notably, it nearly doubles the MobileWorld Pass@3 success rate and yields consistent Pass@1 improvements on AndroidWorld across all three model scales (up to +9.20 points).
\end{itemize}

\section{Related Work}
\label{sec:related-work}

\textbf{Synthetic Data for GUI Agents.}\label{sec:related-synthetic-data} Training GUI agents is bottlenecked by the scarcity of high-quality trajectories, since manual annotation is costly and hard to scale. A first line of work derives trajectories from active environment interaction, using random-walk or exploratory policies to collect step-level actions and state transitions, as in OS-Genesis~\cite{osgenesis} and GUIrilla~\cite{guirilla}, with GUI-ReWalk~\cite{guirewalk} adding reasoning-guided exploration and WebSynthesis~\cite{websynthesis} simulating transitions via an LLM world model with Monte Carlo Tree Search. A second line reuses existing human knowledge artifacts: AgentTrek harvests and filters web tutorials into structured task specifications replayed by a VLM agent~\cite{agenttrek2024}, while Video2GUI mines unlabeled Internet videos to extract grounded interaction trajectories~\cite{video2gui}. A third line trains agents in synthetic environments with verifiable rewards, combining auto-synthesized apps with code-native assertions~\cite{guigenesis2026}, self-supervised pre-training~\cite{uioceanus2026}, few-shot demonstrations~\cite{pcagente2026}, and multi-turn reinforcement learning~\cite{uitars2_2025}. In contrast, Infinite-Dreamer synthesizes pixel-level trajectories by editing screenshots directly, bypassing both environment interaction and reward design.

\textbf{World Models for Visual Environments and GUIs.}
\label{sec:related-world-models}
World models learn environment dynamics to support downstream planning, representation learning, and synthetic rollouts. Beyond classical latent-space formulations~\citep{dreamerv2,dreamerv3}, recent general world models leverage expressive diffusion models to synthesize high-fidelity visual frames for video generation~\citep{worlddreamer2024,genie2024,pandora2024,gamegenx2024,adaworld2025}, video editing~\citep{avid2024}, and diverse interactive environment simulation. Some simulate the environment through text or HTML rendering~\citep{qwenagentworld2026}, which discards the visual dynamics that a screenshot-based agent actually observes. Others train visual simulators from scratch: ViMo~\citep{vimo2025} builds a generative visual world model for mobile app agents, MobileDreamer~\citep{mobiledreamer2026} uses sketch-based abstractions for efficient rollout imagination, and NeuralOS~\citep{neuralos2025} simulates OS-scale dynamics via recurrent state tracking and neural rendering. 

\textbf{Image Editing Models.}
\label{sec:related-image-editing}
Recently, diffusion-based generative models have achieved photorealistic text-to-image synthesis~\citep{stablediffusion2022,dalle3_2023,flux2024,pixart2024,instructpix2pix2022}. Compared with standard text-to-image models, image editing models additionally accept one or more reference images as auxiliary conditioning signals alongside the text prompt, enabling fine-grained controlled modification of existing visuals. This paradigm has rapidly advanced with strong instruction-following editors such as FLUX.1~\citep{flux1kontext2025}, FLUX.2~\citep{flux-2-2025}, and Qwen-Image-Edit~\citep{qwenimageedit2025}. However, we observe that while these models can synthesize photorealistic GUI screenshots, they still lack the underlying procedural knowledge of how apps actually operate within GUI environments, thereby precluding them from serving as genuine GUI world models.

\begin{figure}[tb]
\centering
\includegraphics[width=\textwidth]{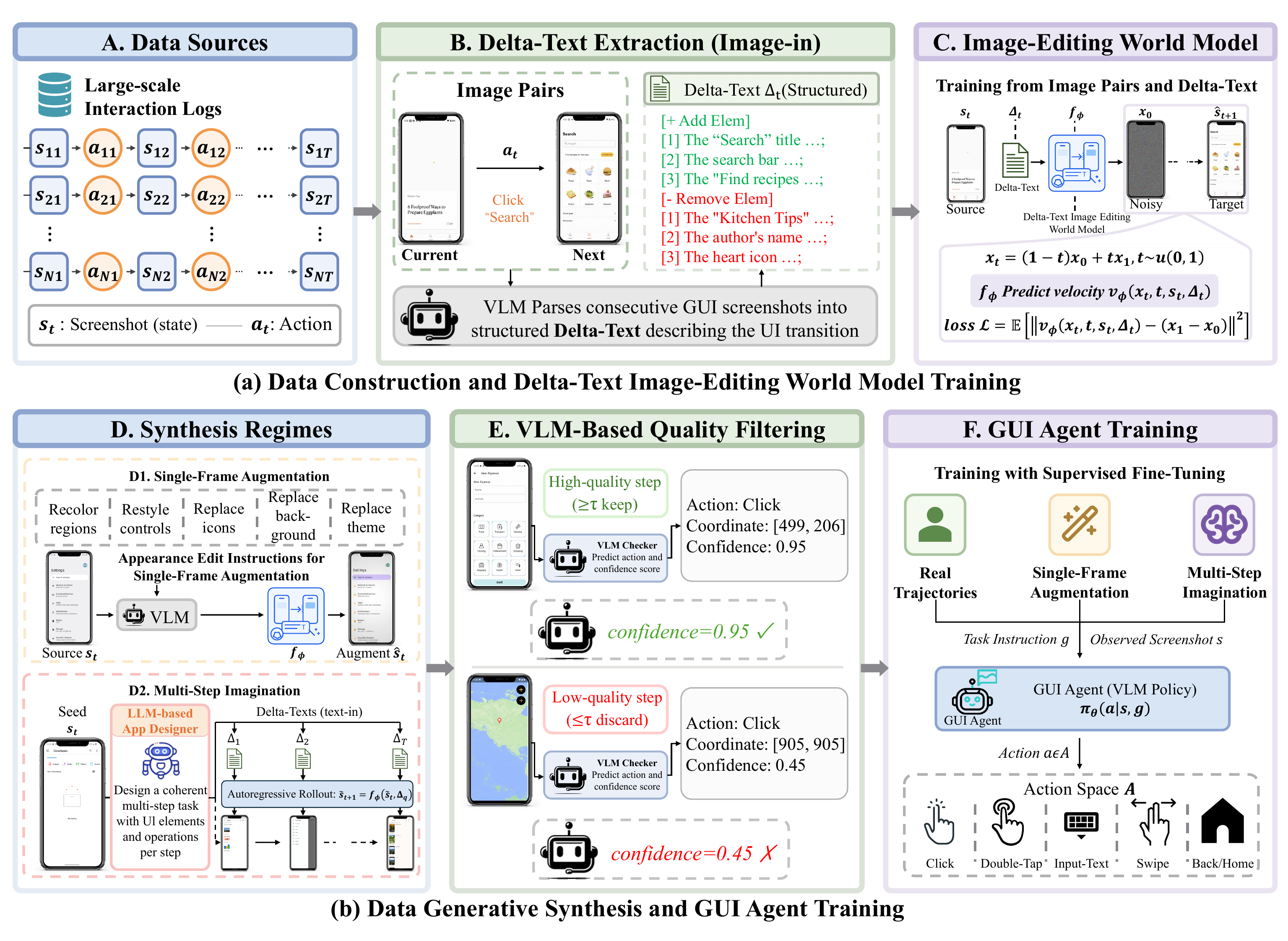}
\caption{\textbf{Overview of our proposed \textit{Infinite-Dreamer}:} (a) Data Construction and Delta-Text Image-Editing World Model Training. We extract structured delta-texts from real UI trajectories and fine-tune an instruction-guided image-editing model as a GUI world model; (b) Data Generative Synthesis and GUI Agent Training. The trained world model enables two synthesis regimes, namely single-frame augmentation and multi-step imagination, whose outputs are filtered by a VLM-based quality checker and used to train the GUI agent.}
\label{fig:framework}
\end{figure}

\section{Methodology}
\label{sec:framework}

\subsection{Problem Formulation}
A mobile GUI agent is considered: at step $t$, the agent observes screenshot $s_t$ and instruction $g$, executes action $a_t$ (e.g., click, type, swipe; Appendix~\ref{sec:appendix-action-space}), and transitions to $s_{t+1} \sim P(\cdot \mid s_t, a_t)$, forming trajectory $\tau = (s_1, a_1, \ldots, a_{T-1}, s_T)$. Collecting real interaction data is bottlenecked by labor costs and exploration dead-ends. We therefore model transition dynamics $P$ with a generative world model to synthesize trajectories $\tilde{\tau}$ for policy learning $\pi_\theta(a \mid s, g)$.

Because raw action labels carry no visual specification, directly predicting $P(s_{t+1} \mid s_t, a_t)$ is ill-posed. We formulate GUI transition as a \emph{hierarchical world model}: $P(s_{t+1} \mid s_t, a_t) = \sum_{\Delta_t} P_\phi(s_{t+1} \mid s_t, \Delta_t) P_\omega(\Delta_t \mid s_t, a_t)$, where $\Delta_t$ is a structured semantic transition specification (\emph{delta-text}). This formulation separates three roles: (i)~the \textbf{App Designer ($G_\omega$)} plans multi-step goals and UI intents; (ii)~the \textbf{delta-text ($\Delta_t$)} specifies the transition and affordance contract; and (iii)~the \textbf{image-editing model ($f_\phi$)} simulates continuous visual dynamics, predicting $\hat{s}_{t+1}$.

\subsection{Delta-Text Extraction}
\label{sec:delta-text}
To build the world model's training set, we use a corpus of real mobile interaction trajectories that provides consecutive (screenshot, action) pairs. For each pair $(s_t, s_{t+1})$ with action $a_t$, a vision-language model extractor $E_\psi$ using a template (Appendix~\ref{sec:appendix-delta-image-prompt}) produces a structured description $\Delta_t$, referred to as the \emph{delta-text}, that characterizes the transition:
\begin{equation}
\Delta_t = E_\psi(s_t, s_{t+1}, a_t, a_{t+1}),
\end{equation}
where $a_{t+1}$ is included so the extractor can identify which UI elements will be relevant for the following step (i.e., the next-step affordances $\mathcal{F}_t$).
Concretely, the output is a record $\Delta_t = (o_t,\, \mathcal{E}_t^{+},\, \mathcal{E}_t^{-},\, \mathcal{F}_t)$ with fields: \textbf{Trigger operation $o_t$:} a natural-language paraphrase of the action $a_t$. \textbf{Added elements $\mathcal{E}_t^{+}$:} UI components that appear or become visible. \textbf{Removed elements $\mathcal{E}_t^{-}$:} UI components that disappear or become hidden. \textbf{Next-step affordances $\mathcal{F}_t$ (optional):} task-relevant elements relevant for the subsequent action. The delta-text thus serves as the world model's unified conditioning interface: it is extracted here from real pairs for training, and later authored from synthetic text-only UI descriptions to faithfully drive trajectory imagination.

\subsection{Image-Editing World Model Training}
We train an image-editing world model $f_\phi$ that predicts the next screenshot $\hat{s}_{t+1}$ conditioned on both the current screenshot $s_t$ and a delta-text $\Delta_t$:
\begin{equation}
\hat{s}_{t+1} = f_\phi(s_t, \Delta_t).
\end{equation}
Starting from a pretrained instruction-guided image-editing backbone, we fine-tune it on the paired screenshots, using $s_t$ as the source image and $\Delta_t$ as the editing instruction. The backbone follows a standard rectified flow-matching formulation. Let $x_0$ be the source image and $x_1$ be the target image; then, with conditioning $c=(s_t,\Delta_t)$ and a velocity network $v_\phi$, we sample $t \sim \mathcal{U}(0,1)$ and minimize the following velocity-prediction objective:
\begin{equation}
\mathcal{L}_{\mathrm{WM}}(\phi) = \mathbb{E}_{t}\!\left[\big\| v_\phi(x_t, t;\, c) - (x_1 - x_0) \big\|_2^2\right],
\end{equation}
where $x_t = (1-t)x_0 + t\, x_1$ linearly interpolates between source and target images. Unlike conventional renderers that need full scene graphs with exact coordinates, $f_\phi$ receives coordinate-free semantic diffs and must infer layout reflow, occlusions, and causal state persistence in unedited regions (Appendix~\ref{sec:appendix-world-model-vs-renderer}); training configurations are in Appendix~\ref{sec:appendix-world-model-training}.

\subsection{Synthesis Regimes}

\paragraph{Single-frame Visual Augmentation.}
Given a screenshot $s$, we generate an appearance-only variant $\tilde{s}$ that edits the visual appearance while preserving the layout and displayed text:
\begin{equation}
\tilde{s} = f_\phi(s, \Delta^{\mathrm{aug}}),
\end{equation}
where $\Delta^{\mathrm{aug}}$ is an appearance-editing delta-text (e.g., recoloring, restyling, or theming; template in Appendix~\ref{sec:appendix-augment-prompt}). To enforce this, the prompt constrains the generation so that $\tilde{s}$ approximately preserves the layout and displayed text of $s$ (i.e., $\mathrm{Layout}(\tilde{s}) \approx \mathrm{Layout}(s)$ and $\mathrm{Text}(\tilde{s}) \approx \mathrm{Text}(s)$), thereby keeping the paired task instruction valid.

\paragraph{Multi-step Imagination with an App Designer.}
To generate high-quality GUI trajectories beyond the coverage of existing data, we introduce an LLM-based App Designer $G_\omega$. Given a seed screenshot $s_1$, $G_\omega$ designs a multi-step task $g$ with per-step UI-element descriptions $\{d_t\}_{t=1}^{T}$ (Appendix~\ref{sec:appendix-trajectories-prompt}). Delta-texts are then authored text-in from these descriptions by a text-only extractor $E_\psi^{\mathrm{txt}}$ (Appendix~\ref{sec:appendix-delta-text-prompt}):
\begin{equation}
\Delta_t = E_\psi^{\mathrm{txt}}(d_t, a_t), \quad t=1,\ldots,T,
\end{equation}
and the world model subsequently rolls out a trajectory autoregressively, starting from the real seed $\tilde{s}_1 = s_1$:
\begin{equation}
\tilde{s}_{t+1} = f_\phi(\tilde{s}_t, \Delta_t), \quad t=1,\ldots,T.
\end{equation}

\paragraph{VLM-based Quality Filtering.}
Since that autoregressive rollouts can accumulate visual artifacts, we thus score every synthesized step with a VLM: it predicts the intended next action's function call together with a confidence $c_t \in [0,100]$. Any step with $c_t < \eta$ is discarded (Appendix~\ref{sec:appendix-relabel}), and the retained steps then form the filtered trajectory:
\begin{equation}
\tilde{\tau}^{\mathrm{filtered}} = \{ (\tilde{s}_t, a_t) : c_t \ge \eta,\; t=1,\ldots,T \}.
\end{equation}

\subsection{GUI Agent Training}
Finally, we fine-tune a VLM-based GUI agent over three complementary data sources:
(i) real trajectories $\mathcal{D}_{\mathrm{real}}$ collected from mobile interactions,
(ii) single-frame augmentations $\mathcal{D}_{\mathrm{aug}}$, and
(iii) quality-filtered multi-step imagination trajectories $\mathcal{D}_{\mathrm{gen}}$.
We minimize a supervised fine-tuning objective over a selected combination of these sources:
\begin{equation}
\mathcal{L}_{\mathrm{mix}}(\theta) = -\sum_{k \in \mathcal{S}} \mathbb{E}_{(s, g, a) \sim \mathcal{D}_k}\big[\log \pi_\theta(a \mid s, g)\big],
\end{equation}
where $\mathcal{S} \subseteq \{\mathrm{real, aug, gen}\}$ denotes the active sources. The resulting agent's architecture and action-selection prompt are described in Appendices~\ref{sec:appendix-decision-loop} and~\ref{sec:appendix-prompt-templates}, and more detailed training configurations are reported in Appendix~\ref{sec:appendix-gui-agent-training}.

\section{Experiment}
\label{sec:experiments}

\newcommand{\inc}{\textcolor{green!55!black}{$\uparrow$}}
\newcommand{\dec}{\textcolor{red!70!black}{$\downarrow$}}

\begin{table}[tb]
\centering
\caption{GUI agent performance evaluation. 
Infinite-Actor (Ours) is trained solely on Infinite-Dreamer synthesis data and paired with its base-model Qwen3-VL counterpart of the same scale; \textcolor{green!55!black}{green $\uparrow$} / \textcolor{red!70!black}{red $\downarrow$} indicates whether Infinite-Actor is higher / lower than its counterpart. The \textbf{best} and \underline{second-best} results in each column are \textbf{bolded} and \underline{underlined}, respectively.}
\label{tab:main-results}
\begingroup
\scriptsize
\renewcommand{\arraystretch}{1.15}
\setlength{\tabcolsep}{4.4pt}
\rowcolors{1}{white}{white}
\begin{tabular}{lcccccccccc}
\toprule
\textbf{Method} & \multicolumn{2}{c}{\textbf{AndroidWorld}} & \multicolumn{2}{c}{\textbf{MobileWorld}} & \multicolumn{3}{c}{\textbf{AC-Curated-Hard}} & \multicolumn{3}{c}{\textbf{AC-Curated-Easy}} \\
\cmidrule(lr){2-3}\cmidrule(lr){4-5}\cmidrule(lr){6-8}\cmidrule(lr){9-11}
& Pass@1 & Pass@3 & Pass@1 & Pass@3 & \mbox{Type Acc} & \mbox{Box Acc} & \mbox{Step SR} & \mbox{Type Acc} & \mbox{Box Acc} & \mbox{Step SR} \\
\midrule
\rowcolor{gray!20}\multicolumn{11}{l}{\textit{General-Purpose Models}} \\
Claude-Sonnet-4.5 & 19.25 & 25.86 & 33.92 & 42.11 & \textbf{73.40} & 22.20 & 30.60 & 80.90 & 22.60 & 35.00 \\
Gemini-3-Flash & 34.91 & 56.03 & \textbf{43.27} & \textbf{52.63} & 73.10 & 48.10 & 46.00 & \textbf{84.40} & 72.40 & 68.20 \\
MiniMax-M3 & 10.34 & 17.24 & 14.03 & 23.68 & 70.90 & 47.40 & 41.60 & 75.00 & 45.60 & 42.50 \\
MiMo-V2.5 & 30.46 & 49.14 & 36.55 & \textbf{52.63} & 73.20 & 63.40 & 53.70 & 81.40 & 77.10 & 68.00 \\
Kimi-K2.6 & 32.18 & 45.69 & \underline{40.94} & \underline{49.12} & 69.00 & 48.60 & 46.50 & 79.30 & 69.20 & 61.20 \\
\midrule
\rowcolor{gray!20}\multicolumn{11}{l}{\textit{GUI-Owl-1.5}} \\
GUI-Owl-1.5-2B & 29.31 & 37.93 & 23.39 & 29.82 & 60.20 & 49.70 & 30.30 & 61.50 & 76.60 & 47.50 \\
GUI-Owl-1.5-4B & 8.05 & 15.52 & 23.68 & 28.07 & 70.90 & 74.50 & 57.70 & 77.50 & \underline{95.30} & 74.70 \\
GUI-Owl-1.5-8B & 23.71 & 36.21 & 32.46 & 38.60 & 66.20 & 71.10 & 52.60 & 69.90 & 91.80 & 65.40 \\
\midrule
\rowcolor{gray!20}\multicolumn{11}{l}{\textit{Base-Model Qwen3-VL vs.\ Infinite-Actor (Ours)}} \\
Qwen3-VL-2B & 22.56 & 29.31 & 5.85 & 8.77 & 63.80 & 56.30 & 38.30 & 76.40 & 92.90 & 72.10 \\
$\hookrightarrow$ Infinite-Actor-2B & 31.61\inc & 49.14\inc & 10.53\inc & 17.54\inc & 59.20\dec & 56.80\inc & 42.30\inc & 66.50\dec & 81.40\dec & 61.10\dec \\
\noalign{\vspace{2pt}}\hdashline\noalign{\vspace{2pt}}
Qwen3-VL-4B & 30.17 & 43.10 & 9.65 & 11.40 & 68.60 & \textbf{77.00} & 57.10 & 81.10 & \textbf{95.50} & 78.60 \\
$\hookrightarrow$ Infinite-Actor-4B & \underline{39.37}\inc & 55.17\inc & 14.91\inc & 19.30\inc & 64.40\dec & 64.70\dec & 44.20\dec & \underline{83.70}\inc & 93.20\dec & \textbf{79.90}\inc \\
\noalign{\vspace{2pt}}\hdashline\noalign{\vspace{2pt}}
Qwen3-VL-8B & \underline{39.37} & \underline{56.90} & 8.77 & 11.40 & 72.30 & \underline{76.90} & \textbf{61.60} & 78.60 & \underline{95.30} & 76.00 \\
$\hookrightarrow$ Infinite-Actor-8B & \textbf{43.82}\inc & \textbf{62.07}\inc & 15.20\inc & 21.93\inc & \underline{73.30}\inc & 71.30\dec & \underline{57.90}\dec & 82.40\inc & 93.80\dec & \underline{78.80}\inc \\
\bottomrule
\end{tabular}%
\endgroup
\end{table}

\subsection{Experimental Setup}

\paragraph{Datasets.}
The world model is fine-tuned on the AndroidControl~\citep{li2024effects} train split and evaluated on its validation split by GLM-4.7v model (using the template in Appendix~\ref{sec:appendix-eval-template}). The GUI Agent Infinite-Actor is trained solely on the Infinite-Dreamer synthesis data (composition in Appendix~\ref{sec:appendix-synthesized-data}), without using any real trajectories. We evaluate our model on AndroidWorld~\citep{androidworld} and MobileWorld~\citep{mobileworld}, two online benchmarks that require completing end-to-end tasks in dynamic mobile environments, and on AndroidControl-Curated-Hard/Easy~\citep{androidcontrolcurated}, an additional offline step-level benchmark.

\paragraph{Baselines.}
For the world model, to evaluate it, we compare five general-purpose image editors (FLUX.1 Kontext, Step1X-Edit-v1.2, LongCat-Image-Edit, Qwen-Image-Edit-2511, FireRed-Image-Edit-1.0) and the un-fine-tuned FLUX.2-Klein backbone (4B/9B), also the starting point for our Infinite-Dreamer-4B/9B; we fine-tune the latter on UI screenshot pairs with delta-text. Importantly, the un-fine-tuned backbone isolates the gain from delta-text fine-tuning, while the general-purpose editors serve as off-the-shelf alternatives. Similarly, for the GUI agent, we compare general-purpose VLMs (Claude-Sonnet-4.5, Gemini-3-Flash, MiniMax-M3, MiMo-V2.5, Kimi-K2.6), GUI-specialized GUI-Owl-1.5 (2B/4B/8B), and base-model Qwen3-VL (2B/4B/8B). Implementation details for all baselines are in Appendix~\ref{sec:appendix-baselines}.

\paragraph{Metrics.}
For world-model synthesis quality, we report four VLM-judged dimensions: Accuracy (Acc.), Completeness (Comp.), Contextual alignment (Cont.), and Visual preservation (Vis.), measuring whether the change is correctly applied, contextually sensible, and leaves unedited regions intact. For the GUI agent, we evaluate from two complementary perspectives: task success rate in dynamic environments and step-level accuracy on static datasets. In AndroidWorld and MobileWorld, we use Pass@1 and Pass@3, the task success rates with one and three attempts, respectively. On the static AndroidControl-Curated (Hard/Easy) dataset, we use Type Acc (action-type accuracy), Box Acc (coordinate grounding accuracy), and Step SR (step-level success rate). Detailed evaluation protocols and world-model inference configurations are in Appendices~\ref{sec:appendix-evaluation-protocols} and~\ref{sec:appendix-inference-performance}.

\begin{table}[tb]
\centering
\caption{World-model synthesis quality. Single-step instruction fidelity on the AndroidControl validation set and multi-step rollout quality, both scored across four VLM-judged dimensions (Accuracy, Completeness, Contextual alignment, Visual preservation). Higher is better.}
\label{tab:synthesis-quality-fidelity}
\label{tab:synthesis-quality-rollout}
\small
\setlength{\tabcolsep}{7.1pt}
\renewcommand{\arraystretch}{1.15}

\begin{tabular}{lccccccc}
\toprule
& \multicolumn{4}{c}{\textbf{Single-Step Fidelity}} & \multicolumn{3}{c}{\textbf{Multi-Step Rollout}} \\
\cmidrule(lr){2-5}\cmidrule(lr){6-8}
\textbf{Method} & \textbf{Acc.} & \textbf{Comp.} & \textbf{Cont.} & \textbf{Vis.} & \textbf{1 Step} & \textbf{5 Steps} & \textbf{10 Steps} \\
\midrule
FLUX.1 Kontext (12B) & 33.33 & 33.01 & 36.05 & 34.34 & 34.18 & 22.16 & 27.69 \\
Step1X-Edit-v1.2 (12B) & 34.27 & 34.11 & 36.85 & 30.60 & 33.96 & 48.95 & 39.61 \\
LongCat-Image-Edit (6B) & 33.79 & 32.66 & 36.34 & 40.65 & 35.86 & 20.89 & 32.69 \\
Qwen-Image-Edit-2511 (20B) & 50.30 & 60.50 & 57.44 & 55.52 & 55.94 & 39.44 & 26.15 \\
FireRed-Image-Edit-1.0 (20B) & \underline{64.34} & \underline{67.73} & \underline{71.24} & 58.89 & \underline{65.55} & 47.46 & \underline{40.96} \\
\midrule
FLUX.2 Klein 4B & 45.13 & 46.02 & 51.21 & 46.02 & 47.09 & 24.96 & 28.65 \\
$\hookrightarrow$ Infinite-Dreamer-4B (Ours) & 60.10 & 64.50 & 67.84 & \underline{60.28} & 63.18 & \underline{54.06} & 40.57 \\
\noalign{\vspace{2pt}}\hdashline\noalign{\vspace{2pt}}
FLUX.2 Klein 9B & 62.44 & 65.17 & 68.95 & 59.43 & 64.00 & 42.53 & 29.61 \\
$\hookrightarrow$ Infinite-Dreamer-9B (Ours) & \textbf{68.13} & \textbf{72.27} & \textbf{75.87} & \textbf{66.55} & \textbf{70.71} & \textbf{64.21} & \textbf{48.84} \\
\bottomrule
\end{tabular}
\end{table}

\begin{figure}[tb]
\centering
\begin{subfigure}{0.485\textwidth}
  \centering
  \includegraphics[width=\linewidth]{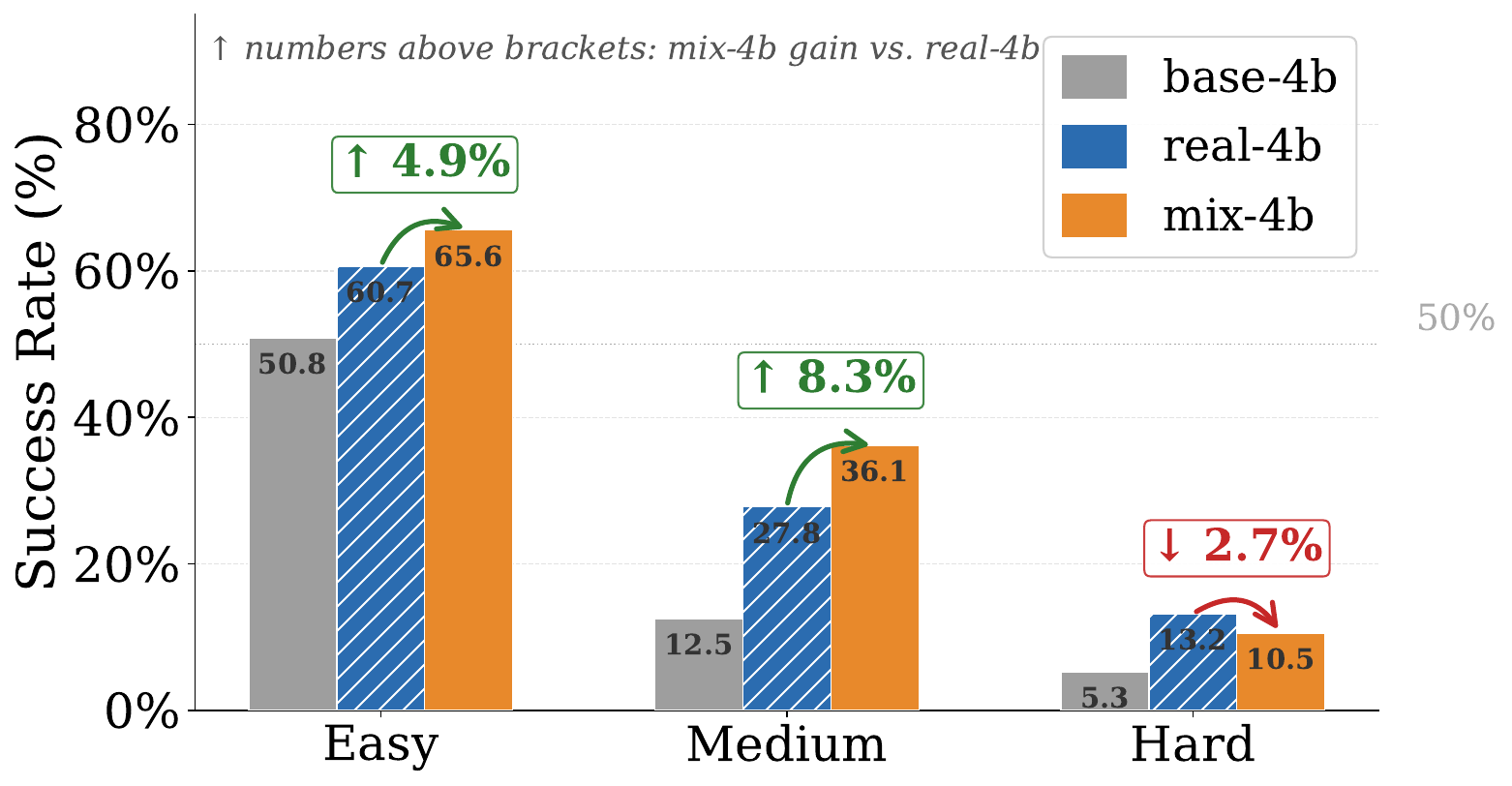}
  \caption{Success rate by task difficulty.}
  \label{fig:difficulty-comparison}
\end{subfigure}\hfill
\begin{subfigure}{0.485\textwidth}
  \centering
  \includegraphics[width=\linewidth]{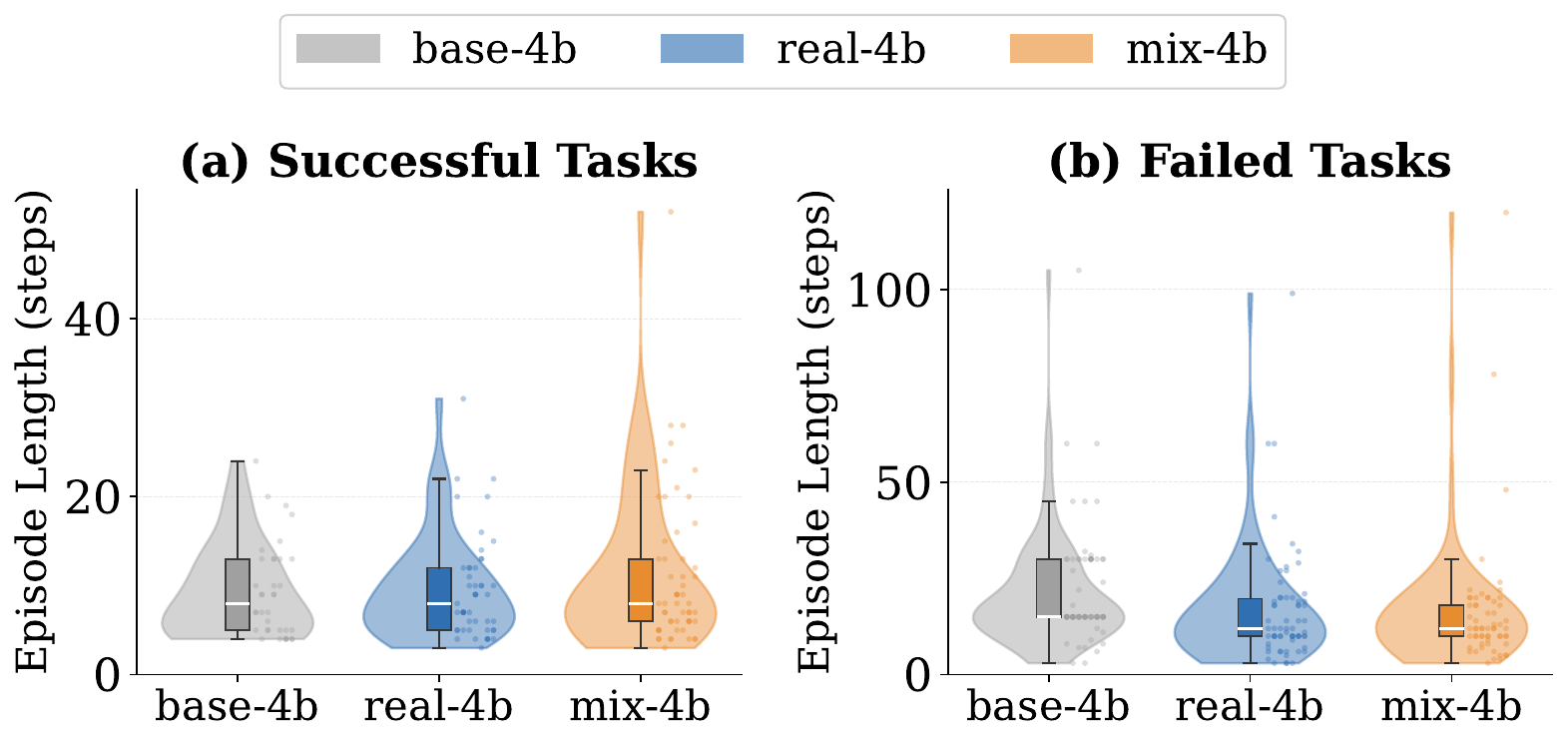}
  \caption{Distribution of trajectory lengths.}
  \label{fig:step-distribution}
\end{subfigure}
\caption{AndroidWorld analyses for base-4b, real-4b, and mix-4b.}
\label{fig:androidworld-analyses}
\end{figure}

\subsection{World Model Quality}

\paragraph{Single-step Synthesis Fidelity.}
As Table~\ref{tab:synthesis-quality-fidelity} shows, Infinite-Dreamer-9B leads on all four metrics, outperforming both the unfine-tuned FLUX.2 Klein 9B backbone and the strongest baseline, FireRed-Image-Edit-1.0. The 4B variant also surpasses its FLUX.2 Klein 4B counterpart, confirming that delta-text-guided fine-tuning on UI data improves instruction fidelity and layout preservation. Qualitative single-step editing comparisons against all baselines are shown in Appendix~\ref{sec:appendix-single-step-comparison}.

\begin{figure}[t!]
\centering
\includegraphics[width=\textwidth]{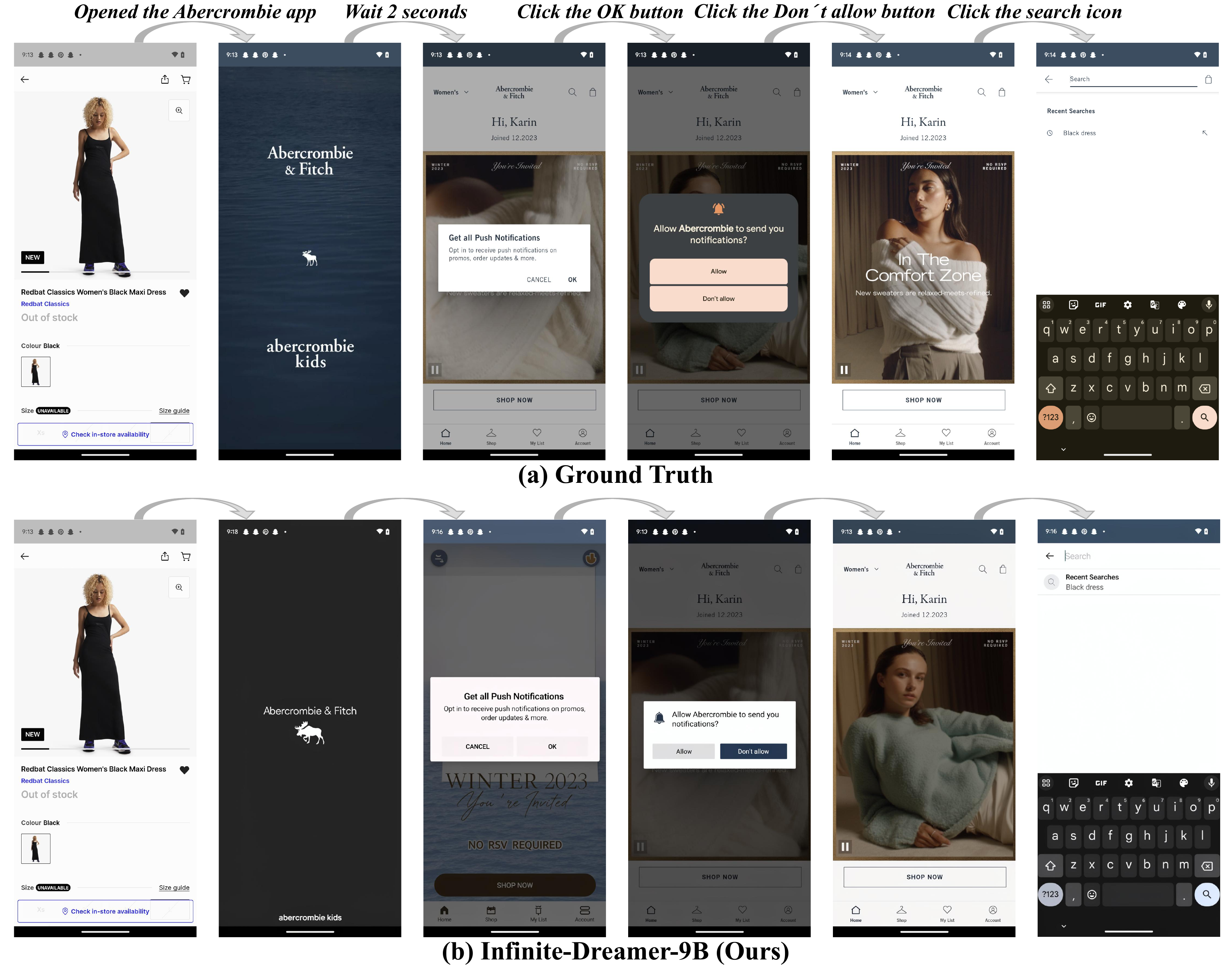}
\caption{Qualitative synthesis comparison. (a) Ground Truth trajectory (top) versus (b) Infinite-Dreamer-9B (Ours) (bottom) over a 6-step GUI interaction sequence (from opening ``Abercrombie'' to clicking the search icon). Our model faithfully reproduces the real UI state transitions, maintaining layout accuracy, text fidelity, and visual coherence across all steps.}
\label{fig:multi-step-synthesis-comparison}
\end{figure}

\paragraph{Multi-step Rollout Stability.}
Table~\ref{tab:synthesis-quality-rollout} shows that Infinite-Dreamer-9B maintains the highest fidelity across all rollout horizons. At 10 steps it scores 48.84, versus 40.96 for FireRed-Image-Edit-1.0 and 29.61 for the unfine-tuned FLUX.2 Klein 9B, demonstrating better stability under recursive application. Figure~\ref{fig:multi-step-synthesis-comparison} shows a representative qualitative multi-step rollout: Infinite-Dreamer-9B faithfully reproduces the ground-truth UI transitions across a 6-step GUI interaction, preserving layout, text content, and visual coherence. Additional qualitative comparisons appear in Appendix~\ref{sec:appendix-multi-step-comparison}. We also discuss failure modes and limitations with examples in Appendix~\ref{sec:appendix-failure-cases}.

\subsection{GUI Agent Performance}

\paragraph{Task Success Rate on AndroidWorld and MobileWorld.}
As Table \ref{tab:main-results} shows, on AndroidWorld, Infinite-Actor-8B achieves 43.82 Pass@1 and 62.07 Pass@3, improving over its base-model counterpart Qwen3-VL-8B (39.37 and 56.90) by +4.45 and +5.17 points, respectively. The gains hold at smaller scales: Infinite-Actor-4B matches the 8B base-model on Pass@1 (39.37) with half the parameters, and Infinite-Actor-2B improves over Qwen3-VL-2B by +9.05 on Pass@1 and +19.83 on Pass@3. On MobileWorld, Infinite-Actor-8B reaches 15.20 Pass@1 and 21.93 Pass@3, against 8.77 and 11.40 for Qwen3-VL-8B base-model, nearly doubling the success rate. These results confirm that Infinite-Dreamer synthesis data provides effective supervision for mobile tasks. More results and analysis are provided in Appendix~\ref{sec:appendix-additional-results}.

\paragraph{Step Level Accuracy on AndroidControl-Curated (AC-Curated).}
Table~\ref{tab:main-results} shows that Infinite-Actor achieves strong action-type accuracy (73.30 on Hard and 82.40 on Easy for 8B) but regresses in coordinate grounding (Box Acc: $77.00 \rightarrow 64.70$ on 4B Hard). The underlying box predictions are displaced rather than marginally noisy: over the $5{,}016$ Easy and $5{,}074$ Hard annotated-box steps, the median distance between predicted and annotated box centers grows from $0.06$ to $0.13$ annotated-box widths on Easy and from $0.18$ to $0.38$ on Hard, with 90th percentiles of $0.83 \rightarrow 0.88$ and $1.90 \rightarrow 4.73$. These errors are not near misses: $88\%$ (Easy) and $97\%$ (Hard) of the model's incorrect boxes place their center more than $50$ pixels outside the annotation. Because the offline metric accepts a box only when its center falls inside the annotation, the loss concentrates on compact targets: on Hard, Box Acc is $46.7\%$ for the smallest third of annotations against $75.4\%$ for the largest third (base: $67.5\%$ vs.\ $87.0\%$). Online, padded touch targets and post-action feedback absorb part of the same displacement, so an offset tap need not end an episode (Appendix~\ref{sec:appendix-failure-cases}).

\subsection{Ablation Study}

\begin{table}[tb]
\centering
\small
\caption{Data source ablation on AndroidWorld Pass@1 (\%), using Qwen3-VL-4B as the fixed backbone. Real Data: real collected trajectories; Gen Data: multi-step imagination from the world model; Aug Data: single-frame visual augmentation.}
\label{tab:ablations}
\setlength{\tabcolsep}{32.2pt}
\renewcommand{\arraystretch}{1.0}

\begin{tabular}{ccc c}
\toprule
\textbf{Real Data} & \textbf{Gen Data} & \textbf{Aug Data} & \textbf{SR (\%)} \\
\midrule
$\times$ & $\times$ & $\times$ & 30.17 \\
$\times$ & \checkmark & \checkmark & 39.37 \\
\checkmark & $\times$ & $\times$ & 42.67 \\
\checkmark & \checkmark & $\times$ & 44.83 \\
\checkmark & $\times$ & \checkmark & 47.41 \\
\rowcolor{gray!20}\checkmark & \checkmark & \checkmark & 48.28 \\
\bottomrule
\end{tabular}

\caption{World model design ablation on instruction fidelity. Both variants use FLUX.2-Klein-9B as the backbone. ``w/o delta-text'' indicates that the world model is conditioned only on the trigger action, completely omitting fine-grained descriptions of added or removed UI elements.}
\label{tab:world-model-ablations}
\setlength{\tabcolsep}{23pt}

\begin{tabular}{lcccc}
\toprule
\textbf{Method} & \textbf{Acc.}  & \textbf{Comp.} & \textbf{Cont.} & \textbf{Vis.} \\
\midrule
\textbf{Infinite-Dreamer-9B} & \textbf{68.13} & \textbf{72.27} & \textbf{75.87} & \textbf{66.55} \\
\multicolumn{1}{r}{\quad w/o delta-text} & 21.56 & 19.72 & 26.14 & 25.57 \\
\bottomrule
\end{tabular}
\end{table}

\paragraph{Synthesis Regimes Analysis.}
We additionally vary the inclusion of real trajectories from OpenMobile~\citep{openmobile} to isolate the contribution of each data source on AndroidWorld. Adding multi-step imagination to real-only training (42.67) lifts SR to 44.83, and adding augmentation instead reaches 47.41; combining all three sources gives the best result, 48.28. These two synthesis regimes therefore contribute in distinct ways: augmentation mainly improves robustness to surface-level UI variation, while multi-step imagination expands the coverage of state transitions. More visualizations appear in Appendices~\ref{sec:appendix-augmentation-results} and~\ref{sec:appendix-imagination-results}.

\paragraph{Performance Across Task Complexities and Hard Tasks.}
Stratifying AndroidWorld by difficulty (Figure~\ref{fig:difficulty-comparison}) shows that mixed training improves Easy (+4.9) and Medium tasks (+8.3), where expanded state coverage aids exploration. On Hard tasks, however, mix-4b underperforms real-4b by -2.7 points (10.5 vs.\ 13.2). Hard tasks demand long, strictly ordered interactions (e.g., cross-application state transfer and precise system dialogs), and world-model fidelity decays over such horizons (Table~\ref{tab:synthesis-quality-rollout}), injecting causal noise that narrower real demonstrations avoid. In the Hard-task logs, 41.2\% of the failures exhaust the step budget and 35.3\% falsely declare completion (Appendix~\ref{sec:appendix-failure-cases}).

\paragraph{Trajectory-length Analysis.}
Figure~\ref{fig:step-distribution} compares AndroidWorld episode lengths by task outcome for base-4b, real-4b, and mix-4b. On successful tasks the three models occupy a similar band of short episodes, and the upper tail separates them: mix-4b's successful episodes run past 50 steps, where real-4b stops near 30 and base-4b near 25. Multi-step synthetic rollouts therefore extend the horizon the policy can sustain. Failed episodes run longer than successful ones for every model, consistent with runs that exhaust the step budget, and mix-4b's failed distribution tracks the baselines.

\paragraph{Necessity of Delta-text Conditioning and Error Dynamics.}
Table~\ref{tab:world-model-ablations} shows that removing delta-text collapses synthesis quality (Acc.\ drops from 68.13 to 21.56), confirming its indispensable role in transition grounding. However, this dependence exposes sensitivity to specification errors: training delta-texts come from real image pairs ($E_\psi$), while rollout delta-texts are authored text-only ($E_\psi^{\mathrm{txt}}$). During imagination, the text-only extractor must infer the diff from a planned UI description without ground-truth visual feedback, so underspecified diffs (e.g., a missing keyboard dismissal or background dimming) can produce visual distortions that compound autoregressively (Table~\ref{tab:synthesis-quality-rollout}). Our VLM quality filter ($\eta=90$) prunes such steps before training (formal error dynamics in Appendix~\ref{sec:appendix-error-propagation}; qualitative ablation in Appendix~\ref{sec:appendix-additional-results}).

\FloatBarrier

\section{Conclusion}
\label{sec:conclusion}

In this paper, we present Infinite-Dreamer, a simulation-free data synthesis framework leveraging an image-editing world model for scalable GUI agent training. By conditioning generation on structured delta-text, our framework controllably synthesizes screenshot transitions and multi-step rollouts. Extensive experiments demonstrate that this synthetic supervision substantially enhances agent generalization, yielding large task-level gains for Infinite-Actor on AndroidWorld and MobileWorld. More broadly, our work highlights a promising direction where generative world models and autonomous agents co-evolve: high-fidelity world models expand state coverage, while agents identify transitions critical for long-horizon competence.

\subsection*{AI use statement}

This paper uses large vision-language models as instruments inside the proposed
method. Concretely, GLM-4.7v serves as the delta-text extractor that produces
the conditioning descriptions (Section~\ref{sec:delta-text}) and as the judge
for the synthesis-quality metrics
(Appendix~\ref{sec:appendix-eval-template}), and Qwen3.5-397B-A17B serves as the
quality filter applied to generated trajectories
(Appendix~\ref{sec:appendix-relabel}). All of these components are described in
the Method and in Appendix~\ref{sec:appendix-experimental-details}, and every
model output that enters the training data is filtered and was reviewed by the
authors.

We take responsibility for the final content of this work, including all text,
claims, and artifacts.

\subsection*{Ethics statement}

This paper studies data synthesis for GUI agent training. All screenshots and
trajectories used in the experiments come from the public AndroidControl,
AndroidWorld, MobileWorld, and AndroidControl-Curated benchmarks. We collect no
new human data and conduct no user studies, so no IRB approval was required.
The synthesized screenshots are derived from these public sources and will be
released under a research-permissive license.

Two risks deserve note. First, synthesized screenshots inherit the distribution
of their source data, so the application coverage and language mix of our
synthesized corpus reflect those of AndroidControl; agents trained on it may
therefore generalize unevenly across regions and applications. Second, a world
model that predicts GUI transitions at pixel level is dual-use, since the same
capability that supports agent training can also automate harmful workflows. We
restrict the study to standard mobile-assistant benchmarks and release only the
synthesized screenshots together with the generation scripts, not any
device-control harness.

\subsection*{Reproducibility statement}

We specify the full experimental configuration in the appendix so that the
results can be reproduced. Appendix~\ref{sec:appendix-experimental-details}
lists the world-model and GUI-agent training hyperparameters and the inference
configuration, and Appendix~\ref{sec:appendix-evaluation-protocols} defines the
evaluation metrics. Every prompt template used in the pipeline is reproduced
verbatim: delta-text extraction
(Appendices~\ref{sec:appendix-delta-image-prompt}
and~\ref{sec:appendix-delta-text-prompt}), single-frame augmentation
(Appendix~\ref{sec:appendix-augment-prompt}), multi-step imagination
(Appendix~\ref{sec:appendix-trajectories-prompt}), quality filtering
(Appendix~\ref{sec:appendix-relabel}), and synthesis-quality evaluation
(Appendix~\ref{sec:appendix-eval-template}).
Appendix~\ref{sec:appendix-reproducibility} records the random seeds, the
number of runs per configuration, and the computing infrastructure.
Code is available at \url{https://github.com/swaydy-n/Infinite-Dreamer}. The synthesized dataset and the generation and
training scripts will be released upon publication.

\bibliographystyle{iclr2027}
\bibliography{refs}

\appendix
\setcounter{secnumdepth}{2}

\section{Experimental Details}
\label{sec:appendix-experimental-details}

This section provides detailed experimental configurations for reproducibility, organized into training settings for the world model and GUI agent, evaluation protocols, and baseline implementations.

\subsection{World Model Training Settings}
\label{sec:appendix-world-model-training}

We build the image-editing world model on FLUX.2-Klein backbones (4B and 9B variants) and fine-tune both on paired UI screenshots with delta-text conditioning. All training configurations are summarized in Table~\ref{tab:world-model-config}.

\begin{table}[h]
\centering
\caption{World model training configuration. Both 4B and 9B models share the same hyperparameters unless otherwise noted.}
\label{tab:world-model-config}
\small
\setlength{\tabcolsep}{30pt}
\begin{tabular}{ll}
\toprule
\textbf{Configuration} & \textbf{Value} \\
\midrule
Backbone & FLUX.2-Klein (4B / 9B) \\
Training data & AndroidControl train split \\
Fine-tuning method & LoRA \\
LoRA rank & 32 \\
Training epochs & 3 \\
Batch size & 1 \\
Learning rate & $1 \times 10^{-4}$ \\
LR schedule & Cosine decay \\
Warmup steps & N/A \\
Optimizer & AdamW \\
Weight decay & 0.01 \\
Resolution & $672 \times 1536$ \\
Precision & bfloat16 \\
Hardware & 4$\times$ NVIDIA A100 80GB \\
Random seed & 42 \\
Software & PyTorch 2.9.1 \\
\bottomrule
\end{tabular}
\end{table}

\subsection{Inference Performance of World Models}
\label{sec:appendix-inference-performance}

We evaluate the inference performance of the FLUX.2-Klein 4B and 9B variants as image-editing world models. The evaluation is conducted under the configuration summarized in Table~\ref{tab:inference-config}. The inference performance results are reported in Table~\ref{tab:inference-performance}.

\begin{table}[h]
\centering
\caption{Inference configuration for performance testing.}
\label{tab:inference-config}
\small
\setlength{\tabcolsep}{20pt}
\begin{tabular}{ll}
\toprule
\textbf{Configuration Item} & \textbf{Value} \\
\midrule
Image Width & 672 px \\
Image Height & 1536 px \\
Inference Steps & 40 steps \\
Guidance Scale & 4.0 \\
\bottomrule
\end{tabular}
\end{table}

\begin{table}[h]
\centering
\caption{Inference performance comparison between FLUX.2-Klein 9B and 4B models.}
\label{tab:inference-performance}
\small
\setlength{\tabcolsep}{15pt}
\begin{tabular}{lrr}
\toprule
\textbf{Metric} & \textbf{Infinite-Dreamer-9B} & \textbf{Infinite-Dreamer-4B} \\
\midrule
Average Time per Image & 57.202 s & 28.779 s \\
Average Time per Step & 1430.0 ms & 719.5 ms \\
Throughput (Img/s) & 0.01748 & 0.03475 \\
Pixel Throughput (Mpx/s) & 0.01804 & 0.03587 \\
\bottomrule
\end{tabular}
\end{table}

\subsection{GUI Agent Training Settings}
\label{sec:appendix-gui-agent-training}

We fine-tune Infinite-Actor agents from Qwen3-VL-2B/4B/8B checkpoints solely on synthesized data, namely single-frame augmentations and quality-filtered multi-step imagination trajectories, without using any real trajectories. All training configurations are summarized in Table~\ref{tab:gui-agent-config}.

\begin{table}[h]
\centering
\caption{GUI agent training configuration. All model sizes (2B/4B/8B) share the same hyperparameters.}
\label{tab:gui-agent-config}
\small
\setlength{\tabcolsep}{30pt}
\begin{tabular}{ll}
\toprule
\textbf{Configuration} & \textbf{Value} \\
\midrule
Backbone & Qwen3-VL (2B / 4B / 8B) \\
Training data & Infinite-Dreamer synthesis data (aug, gen) \\
Fine-tuning method & Full fine-tuning \\
Training steps & 1,000 \\
Batch size & 32 \\
Learning rate & $1.0 \times 10^{-5}$ \\
LR schedule & Cosine decay \\
Warmup steps & 30 \\
Optimizer & AdamW \\
Weight decay & 0.05 \\
Resolution & $672 \times 1536$ \\
Precision & bfloat16 \\
Hardware & 4$\times$ NVIDIA A40 48GB \\
Random seed & 42 \\
Software & PyTorch 2.9.1, transformers 4.57.6 \\
\bottomrule
\end{tabular}
\end{table}

\begin{figure}[h]
\centering
\includegraphics[width=0.6\textwidth]{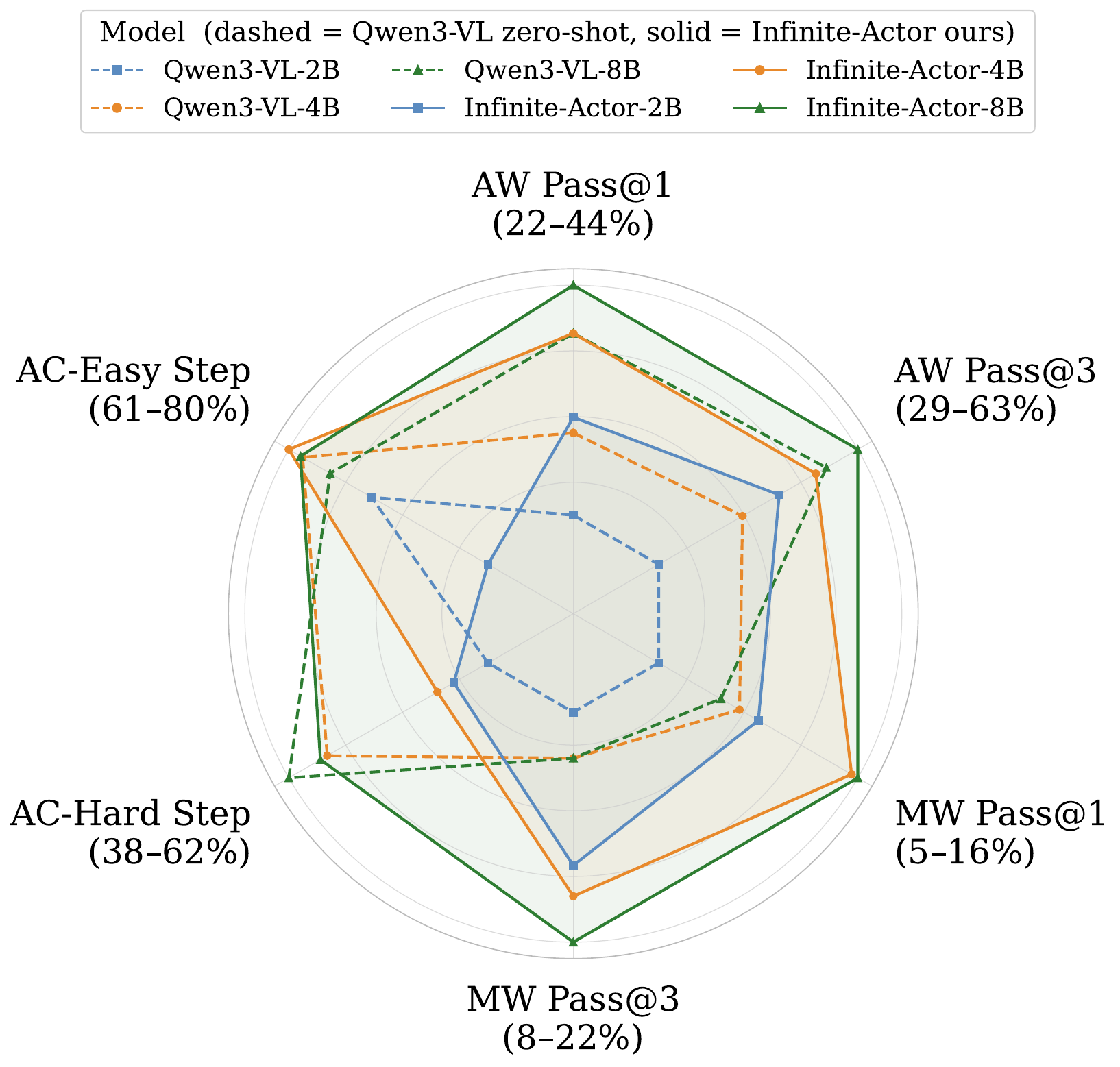}
\caption{Performance comparison between zero-shot Qwen3-VL and Infinite-Actor agents across model scales (2B/4B/8B) on six evaluation metrics from Table~\ref{tab:main-results}: AW Pass@1 (AndroidWorld Pass@1), AW Pass@3, MW Pass@1 (MobileWorld Pass@1), MW Pass@3, AC-Easy Step SR (AC-Curated-Easy Step SR), and AC-Hard Step SR (AC-Curated-Hard Step SR). Infinite-Actor models consistently outperform their zero-shot counterparts on task-level benchmarks.}
\label{fig:appendix-radar-comparison}
\end{figure}

\subsection{Reproducibility Details}
\label{sec:appendix-reproducibility}

\paragraph{Hyperparameter selection.}
We did not conduct an extensive hyperparameter search; a single setting per model was used, selected by preliminary validation on the AndroidControl validation split. The final values are listed in Tables~\ref{tab:world-model-config} and~\ref{tab:gui-agent-config}; all other parameters follow the backbone defaults.

\paragraph{Random seeds.}
All data generation, training, and evaluation procedures use a fixed random seed (seed $=$ 42) to ensure reproducibility.

\paragraph{Algorithm runs.}
All reported results are obtained from a single training run per model configuration; Pass@3 results are computed over three independent inference attempts, as described in Appendix~\ref{sec:appendix-evaluation-protocols}.

\paragraph{Computing infrastructure.}
All experiments are run on Ubuntu 22.04 LTS with Python 3.11. The world model is implemented in PyTorch 2.9.1; the GUI agent is fine-tuned with PyTorch 2.9.1 and transformers 4.57.6. The pipeline VLMs (delta-text extractor, quality filter, and synthesis-quality judge) are served via vLLM 0.16.0. GPU models and memory for each stage are listed in Tables~\ref{tab:world-model-config} and~\ref{tab:gui-agent-config}.

\paragraph{Preprocessing pipeline.}
The complete data preprocessing pipeline---delta-text extraction (Appendix~\ref{sec:appendix-delta-image-prompt} and~\ref{sec:appendix-delta-text-prompt}), single-frame augmentation (Appendix~\ref{sec:appendix-augment-prompt}), multi-step imagination (Appendix~\ref{sec:appendix-trajectories-prompt}), and quality filtering (Appendix~\ref{sec:appendix-relabel})---is fully specified by the prompt templates and configuration tables included in this appendix.

\subsection{Synthesized Data}
\label{sec:appendix-synthesized-data}

The novel data introduced by this paper is the Infinite-Dreamer synthesis set used to train the GUI agent (Section~\ref{sec:framework}, GUI Agent Training), composed of two complementary subsets:
\begin{itemize}
    \item \textbf{Single-frame augmentations $\mathcal{D}_{\mathrm{aug}}$:} appearance-only variants of real screenshots (theme, color, and icon restyling) that preserve layout and text, generated by the world model under the augmentation delta-text (Appendix~\ref{sec:appendix-augment-prompt}).
    \item \textbf{Multi-step imagination trajectories $\mathcal{D}_{\mathrm{gen}}$:} autoregressive rollouts produced by the App Designer and the world model (Appendix~\ref{sec:appendix-trajectories-prompt}), retained step-by-step after VLM quality filtering with confidence threshold $\eta=90$ (Appendix~\ref{sec:appendix-relabel}).
\end{itemize}
Each sample is stored as a $(s_t, g, a_t)$ triple---a screenshot, the task instruction, and a \texttt{mobile\_use} action---at resolution $672\times1536$. Both subsets are regenerated deterministically from the prompt templates and configuration tables included in this appendix, and the full dataset together with its generation scripts will be released under a research-permissive license upon publication.

\subsection{Evaluation Protocols}
\label{sec:appendix-evaluation-protocols}

\paragraph{Task-level benchmarks.} For AndroidWorld and MobileWorld, we report Pass@1 and Pass@3 metrics following the standard evaluation protocol. Each task executes up to the maximum allowed steps. For Pass@3, we run three independent attempts and consider a task successful if any attempt completes correctly.

\paragraph{Step-level benchmarks.} For AndroidControl-Curated (Hard/Easy), we report Type Acc (action-type accuracy), Box Acc (coordinate grounding accuracy), and Step SR (step-level success rate) computed over all annotated steps. Box Acc is computed over the steps whose ground-truth action requires a coordinate (click and long press); a predicted box counts as correct when its center falls inside the ground-truth box, after rescaling the model's $1000\times 1000$ output to the screenshot resolution. Type Acc and Step SR are computed over all steps, and steps whose ground-truth action requires no coordinate are scored on action type alone.

\subsection{Baseline Implementations}
\label{sec:appendix-baselines}

\paragraph{Proprietary models.} We use official API endpoints for Claude-Sonnet-4.5 (Anthropic), Gemini-3-Flash (Google), MiniMax-M3, MiMo-V2.5, and Kimi-K2.6.

\paragraph{Open-source VLMs.} For Qwen3-VL and GUI-Owl-1.5, we use publicly released checkpoints from Hugging Face with the same prompt template as Infinite-Actor agents.

\paragraph{Image-editing baselines.} For baselines in Table~\ref{tab:synthesis-quality-fidelity} (FLUX.1 Kontext, Step1X-Edit-v1.2, LongCat-Image-Edit, Qwen-Image-Edit-2511, FireRed-Image-Edit-1.0), we use default inference settings from official implementations with the same delta-text prompts as Infinite-Dreamer for fair comparison.

\subsection{Additional Experimental Results}
\label{sec:appendix-additional-results}

\begin{figure}[bt]
\centering
\includegraphics[width=\textwidth]{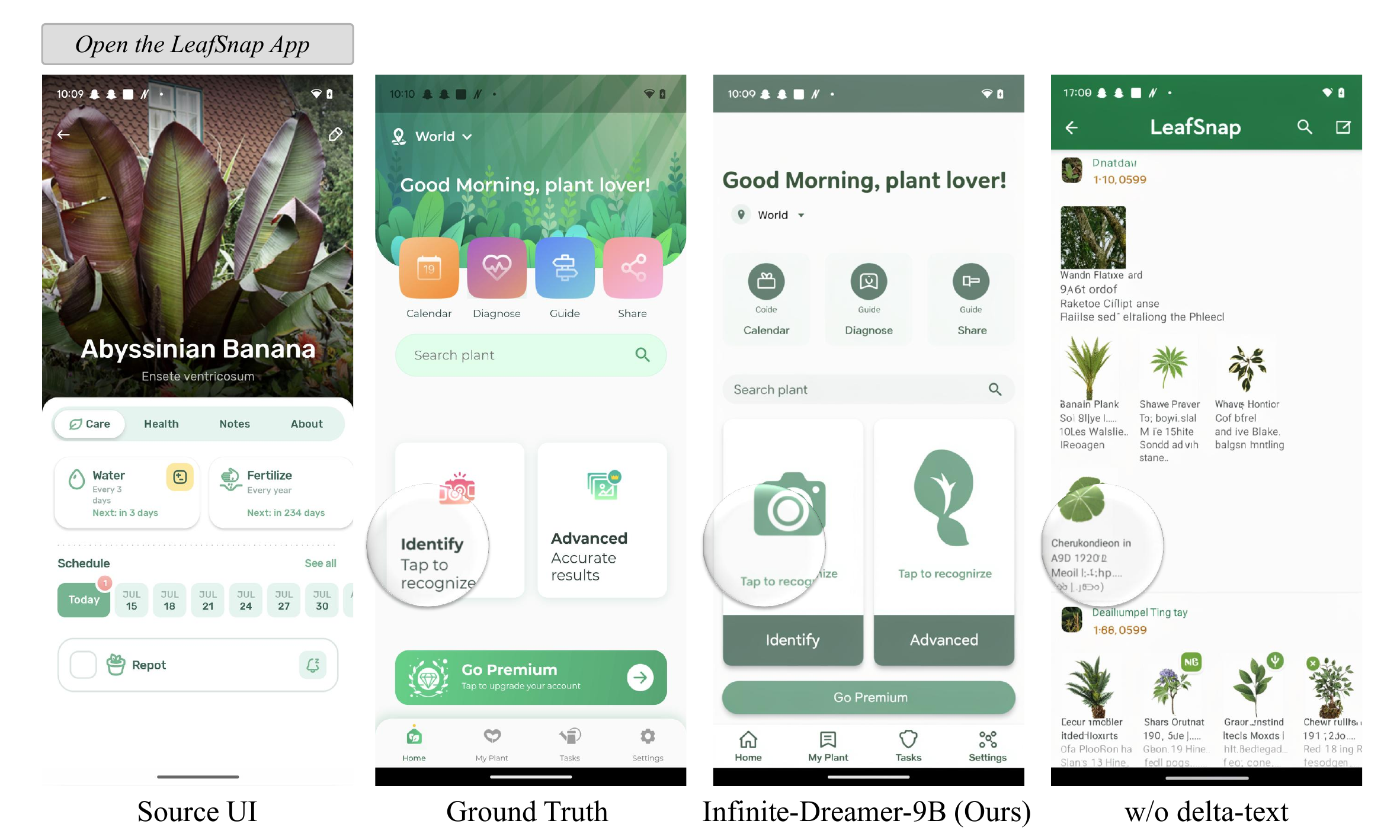}
\caption{Qualitative effect of delta-text conditioning. Given the source UI (leftmost) and the delta-text action ``Open LeafSnap App'', Infinite-Dreamer-9B (third) closely matches the ground-truth transition (second), whereas the variant without delta-text (rightmost, which only receives the trigger action/operation without any fine-grained descriptions of added or removed elements) collapses into an unrelated screenshot that diverges substantially from the intended target UI.}
\label{fig:appendix-delta-text-ablation}
\end{figure}

\paragraph{Qualitative Ablation: Effect of Delta-Text Conditioning.}
Figure~\ref{fig:appendix-delta-text-ablation} illustrates the qualitative effect of delta-text conditioning on a representative example. Given the source UI (leftmost) and the delta-text action ``Open LeafSnap App'', Infinite-Dreamer-9B (third column) closely matches the ground-truth transition (second column), whereas the variant trained without delta-text (rightmost) collapses into an unrelated screenshot that diverges substantially from the intended target UI. This confirms the quantitative finding reported in Table~\ref{tab:world-model-ablations}.

\paragraph{Comparison: Zero-Shot vs.\ Ours.}
Figure~\ref{fig:appendix-radar-comparison} provides a visual summary of the performance gap between zero-shot Qwen3-VL and Infinite-Actor across model scales (2B/4B/8B) on the six evaluation metrics reported in Table~\ref{tab:main-results}: AW Pass@1, AW Pass@3, MW Pass@1, MW Pass@3, AC-Easy Step SR, and AC-Hard Step SR. Infinite-Actor models consistently outperform their zero-shot counterparts on task-level benchmarks.

\FloatBarrier

\subsection{Conceptual Disentanglement: World Model vs.\ Conditional GUI Renderer}
\label{sec:appendix-world-model-vs-renderer}

A central conceptual question is whether our image-editing model acts merely as a conditional graphic renderer rather than a genuine world model. We systematically distinguish between these two paradigms across operational mechanisms, representations, and functional roles in agent training, summarized in Table~\ref{tab:wm-vs-renderer}.

\begin{table}[h]
\centering
\caption{Conceptual and structural comparison between conventional GUI renderers and the Infinite-Dreamer image-editing world model.}
\label{tab:wm-vs-renderer}
\small
\setlength{\tabcolsep}{3.5pt}
\renewcommand{\arraystretch}{1.15}
\begin{tabular}{p{2.6cm}p{5.0cm}p{5.1cm}}
\toprule
\textbf{Dimension} & \textbf{Conventional GUI Renderer (e.g., WebKit)} & \textbf{Infinite-Dreamer World Model ($f_\phi$)} \\
\midrule
\textbf{Input Conditioning} & Exhaustive scene graph (DOM tree, precise $(x,y,w,h)$ coordinates, CSS style properties, typography). & Source screenshot $s_t$ and abstract delta-text $\Delta_t$ without spatial coordinates or component sizes. \\
\midrule
\textbf{Spatial Physics \& Layout} & Explicit, hard-coded geometry calculations and constraint-solver passes. & Implicit visual physics: learned neural inference of layout reflow, occlusions, background dimming, and responsive alignment. \\
\midrule
\textbf{State Persistence} & Redraws the canvas or invalidates explicit tree nodes; each render pass is independent of the previous frame. & Causal state persistence: conditioning on $s_t$ keeps unaffected regions (status bars, background cards) visually consistent with the previous frame. \\
\midrule
\textbf{Affordance Modeling} & Purely passive visual output for human display; oblivious to action viability. & Active affordance preservation: synthesizes interactive visual hot-spots enabling subsequent policy execution. \\
\midrule
\textbf{Operational Role} & Open-loop visual presentation terminal. & Closed-loop state transition simulator $P(s_{t+1} \mid s_t, a_t)$ for trajectory imagination. \\
\bottomrule
\end{tabular}
\end{table}

\paragraph{1. Coordinate-Free Spatial Generalization.}
Conventional graphical rendering pipelines (e.g., Blink, Gecko, or the Android ViewServer) are deterministic rasterization engines. They require complete, unambiguous specifications of every visual element, including exact screen coordinates, width, height, margin, padding, border radius, and color codes. In stark contrast, delta-text conditioning provides only high-level semantic intent (e.g., ``Added: search results dropdown; Removed: search suggestion pill''). The image-editing model $f_\phi$ must infer where and how to render these elements: it predicts valid spatial coordinates, typography, responsive element reflows, and background dimming purely from visual context. This requires internalizing the underlying spatial grammar and visual physics of digital software.

\paragraph{2. Causal State Persistence and Invariance.}
A critical requirement of an environmental world model is state persistence: when an action affects a localized region of the screen, the rest of the environment must remain causally invariant. A conventional renderer treats every render pass as an evaluation of the full DOM. In contrast, $f_\phi$ conditions on the prior visual observation $s_t$ and updates only causally relevant regions, so UI areas unrelated to the action (e.g., device status bars, system navigation icons, background cards) stay visually consistent with the previous frame.

\paragraph{3. Hierarchical Decoupling in World Modeling.}
Rather than conflating high-level task logic with low-level pixel synthesis, our framework adopts a hierarchical world modeling formulation:
\begin{equation}
T = T_{\mathrm{vis}} \circ T_{\mathrm{sem}},
\end{equation}
where $T_{\mathrm{sem}}: (s_t, a_t) \mapsto \Delta_t$ models the symbolic transition dynamics (governed by the App Designer $G_\omega$), and $T_{\mathrm{vis}}: (s_t, \Delta_t) \mapsto s_{t+1}$ models the visual dynamics (governed by $f_\phi$). This hierarchical decoupling mirrors standard world-model architectures in robotics and game simulation, where high-level semantic transition intent is explicitly separated from continuous visual realization to prevent catastrophic visual hallucinations.

\section{GUI Agent Architecture}
\label{sec:appendix-gui-agent}

This section describes the architecture of our mobile GUI agent. The agent interacts with an Android device by observing screenshots, selecting actions from a unified action space (Appendix~\ref{sec:appendix-action-space}), and executing them via ADB (Android Debug Bridge). At each step it receives the current screenshot, the user's goal, and a running task-progress history, selects a single action, and appends the new screenshot together with an updated task-progress entry to the trajectory for the next step; no separate outcome-summarization phase is required.

\subsection{Action Space Definition}
\label{sec:appendix-action-space}

We define a unified action space for mobile GUI interaction that covers common touchscreen operations. Table~\ref{tab:action-space} summarizes all 15 action types and their associated parameters.

\begin{table}[h]
\centering
\caption{Action space for mobile GUI agents. Each action type is listed with its required and optional parameters. Coordinates are specified as $(x, y)$ pixel positions on a $1000 \times 1000$ screen.}
\label{tab:action-space}
\small
\begin{tabular}{@{}llp{5.4cm}@{}}
\toprule
\textbf{Action Type} & \textbf{Parameters} & \textbf{Description} \\
\midrule
\multicolumn{3}{@{}l}{\textit{Task Control}} \\
\texttt{status} & \texttt{goal\_status}\textsuperscript{R} & Report task completion. \texttt{goal\_status} $\in$ \{\texttt{complete}, \texttt{infeasible}\}. \\
\texttt{answer} & \texttt{text}\textsuperscript{R} & Provide a textual answer to a query. \\
\texttt{wait} & --- & Wait for the UI to update or an app to load. \\
\midrule
\multicolumn{3}{@{}l}{\textit{Touch Interactions}} \\
\texttt{click} & \texttt{coordinate}\textsuperscript{R} & Single tap at $(x, y)$. \\
\texttt{double\_tap} & \texttt{coordinate}\textsuperscript{R} & Double tap at $(x, y)$. \\
\texttt{long\_press} & \texttt{coordinate}\textsuperscript{R} & Long press at $(x, y)$ to trigger context menus. \\
\texttt{drag\_and\_drop} & \texttt{coordinate}\textsuperscript{R}, \texttt{coordinate2}\textsuperscript{R} & Drag from $(x_1, y_1)$ to $(x_2, y_2)$. \\
\midrule
\multicolumn{3}{@{}l}{\textit{Text Input}} \\
\texttt{input\_text} & \texttt{text}\textsuperscript{R}, \texttt{coordinate}\textsuperscript{O} & Type text; optionally tap \texttt{coordinate} to focus first. \\
\texttt{keyboard\_enter} & --- & Press the Enter/Return key. \\
\texttt{keyboard\_delete} & --- & Press the Backspace/Delete key. \\
\midrule
\multicolumn{3}{@{}l}{\textit{Navigation}} \\
\texttt{navigate\_home} & --- & Press the Home button. \\
\texttt{navigate\_back} & --- & Press the Back button. \\
\texttt{open\_app} & \texttt{app\_name}\textsuperscript{R} & Launch an application by name. \\
\midrule
\multicolumn{3}{@{}l}{\textit{Scrolling \& Swiping}} \\
\texttt{scroll} & \texttt{direction}\textsuperscript{R} & Scroll the page. \texttt{direction} $\in$ \{\texttt{up}, \texttt{down}, \texttt{left}, \texttt{right}\}. \\
\texttt{swipe} & \texttt{direction}\textsuperscript{R}, \texttt{coordinate}\textsuperscript{O} & Swipe gesture; optionally from a specific point. \\
\bottomrule
\end{tabular}

{\footnotesize \textsuperscript{R}Required parameter. \textsuperscript{O}Optional parameter.}
\end{table}

Each action is invoked via the \texttt{mobile\_use} function with a required \texttt{action\_type} field and action-specific optional parameters. At deployment time the agent acts through a compact \texttt{mobile\_use} tool-call interface (Appendix~\ref{sec:appendix-prompt-templates}) with eight actions; several raw actions are renamed or merged into this vocabulary---for example \texttt{input\_text} as \texttt{type}, \texttt{navigate\_home}/\texttt{navigate\_back}/\texttt{keyboard\_enter} as \texttt{system\_button}, \texttt{scroll}/\texttt{drag\_and\_drop} as \texttt{swipe}, and \texttt{status} as \texttt{terminate}---so the policy emits a single JSON tool call per step.

\subsection{Action-Selection Decision Loop}
\label{sec:appendix-decision-loop}

At each step the agent receives the current screenshot, the user's goal, and a task-progress string summarizing the actions taken so far. The vision-language model outputs a brief Thought explaining its reasoning, a short Action description, and a single JSON \texttt{mobile\_use} tool call specifying the action type and parameters. After parsing the model output, coordinates are normalized from the model's input resolution to the device's native resolution; the action is then executed on the device via ADB, and the agent waits briefly for the UI to stabilize. The newly captured screenshot together with an updated task-progress entry is appended to the trajectory and provided as context for the next action-selection step. This closed loop lets the agent maintain situational awareness across multi-step tasks directly from its action history.

\subsection{Action-Selection Prompt Template}
\label{sec:appendix-prompt-templates}

The complete action-selection prompt is shown below. It defines the \texttt{mobile\_use} tool with eight compact actions (\texttt{click}, \texttt{long\_press}, \texttt{swipe}, \texttt{type}, \texttt{answer}, \texttt{system\_button}, \texttt{wait}, \texttt{terminate}), fixes the output order as Thought $\to$ Action $\to$ \texttt{<tool\_call>}, and injects the user query \texttt{\{\{ query \}\}}, the running task progress \texttt{\{\{ task\_progress \}\}}, and a \texttt{<image>} token for the current screenshot.

\begin{tcolorbox}[%
  title={Action-Selection Prompt},
  colback=white,
  colframe=black!80,
  coltitle=white,
  fonttitle=\bfseries\small,
  boxrule=0.4pt,
  arc=0pt,
  outer arc=0pt,
  left=2mm, right=2mm, top=1mm, bottom=1mm,
  width=\textwidth,
  breakable
]%
\begin{lstlisting}[breaklines=true,basicstyle=\footnotesize\ttfamily,columns=fullflexible,keepspaces=true,literate={–}{{\textendash}}1 {—}{{\textemdash}}1 {‘}{{\textquoteleft}}1 {’}{{\textquoteright}}1 {“}{{\textquotedblleft}}1 {”}{{\textquotedblright}}1]
# Tools

You may call one or more functions to assist with the user query.

You are provided with function signatures within <tools></tools> XML tags:
<tools>
{"type": "function", "function": {"name": "mobile_use", "description": "Use a
touchscreen to interact with a mobile device, and take screenshots.\n* This is an
interface to a mobile device with touchscreen. You can perform actions like
clicking, typing, swiping, etc.\n* Some applications may take time to start or
process actions, so you may need to wait and take successive screenshots to see the
results of your actions.\n* The screen's resolution is 999x999.\n* Make sure to
click any buttons, links, icons, etc with the cursor tip in the center of the
element. Don't click boxes on their edges unless asked.", "parameters":
{"properties": {"action": {"description": "The action to perform. The available
actions are:\n* `click`: Click the point on the screen with coordinate (x, y).\n*
`long_press`: Press the point on the screen with coordinate (x, y) for specified
seconds.\n* `swipe`: Swipe from the starting point with coordinate (x, y) to the
end point with coordinates2 (x2, y2).\n* `type`: Input the specified text into the
activated input box.\n* `answer`: Output the answer.\n* `system_button`: Press the
system button.\n* `wait`: Wait specified seconds for the change to happen.\n*
`terminate`: Terminate the current task and report its completion status.", "enum":
["click", "long_press", "swipe", "type", "answer", "system_button", "wait",
"terminate"], "type": "string"}, "coordinate": {"description": "(x, y): The x
(pixels from the left edge) and y (pixels from the top edge) coordinates to move
the mouse to. The coordinates should be values from 0 to 999. Required only by
`action=click`, `action=long_press`, and `action=swipe`.", "type": "array"},
"coordinate2": {"description": "(x, y): The x (pixels from the left edge) and y
(pixels from the top edge) coordinates to move the mouse to. Required only by
`action=swipe`.", "type": "array"}, "text": {"description": "Required only by
`action=type` and `action=answer`.", "type": "string"}, "time": {"description":
"The seconds to wait. Required only by `action=long_press` and `action=wait`.",
"type": "number"}, "button": {"description": "Back means returning to the previous
interface, Home means returning to the desktop, Menu means opening the application
background menu, and Enter means pressing the enter. Required only by
`action=system_button`", "enum": ["Back", "Home", "Menu", "Enter"], "type":
"string"}, "status": {"description": "The status of the task. Required only by
`action=terminate`.", "type": "string", "enum": ["success", "failure"]}},
"required": ["action"], "type": "object"}}}
</tools>

For each function call, return a json object with function name and arguments
within <tool_call></tool_call> XML tags:
<tool_call>
{"name": <function-name>, "arguments": <args-json-object>}
</tool_call>

# Response format

Response format for every step:
1) Thought: one concise sentence explaining the next move (no multi-step
reasoning).
2) Action: a short imperative describing what to do in the UI.
3) A single <tool_call>...</tool_call> block containing only the JSON: {"name":
<function-name>, "arguments": <args-json-object>}.

Rules:
- Output exactly in the order: Thought, Action, <tool_call>.
- Be brief: one sentence for Thought, one for Action.
- Do not output anything else outside those three parts.
- If finishing, use action=terminate in the tool call.

The user query: {{ query }}.
Task progress (You have done the following operation on the current device): {{
task_progress }}.
<image>\end{lstlisting}
\end{tcolorbox}

\section{Delta-Text Extraction Templates}\label{sec:appendix-diff-prompt}

We extract a structured delta-text $\Delta_t$ for each screenshot transition with one of two templates, chosen by the downstream use of the resulting data. Both produce the same core fields---a trigger operation, added elements, reduced (removed) elements, and the elements required for the next operation---but differ in their inputs and in how much of the surrounding UI they re-state. The image-in variant consumes the before/after screenshots and is used to build the world model's training set, whereas the text-in variant consumes textual UI-element descriptions and is used to build the agent's imagined training data. The full templates are included below.

\subsection{Image-In Extraction for World-Model Training}\label{sec:appendix-delta-image-prompt}

Given two consecutive screenshots (\texttt{<image><image>}) together with the current action \texttt{\{curr\_action\}} and the next action \texttt{\{next\_action\}}, this template asks the vision-language model to describe how the operation changes the page. In addition to the core delta fields it re-states the current and next page elements, yielding six output fields. The resulting delta-texts are paired with the screenshot pairs to construct the training set of the image-editing world model (see the Method section).

\begin{tcolorbox}[%
  title={Image-In Delta-Text Extraction Prompt},
  colback=white,
  colframe=black!80,
  coltitle=white,
  fonttitle=\bfseries\small,
  boxrule=0.4pt,
  arc=0pt,
  outer arc=0pt,
  left=2mm, right=2mm, top=1mm, bottom=1mm,
  width=\textwidth,
  breakable
]%
\begin{lstlisting}[breaklines=true,basicstyle=\footnotesize\ttfamily,columns=fullflexible,keepspaces=true,literate={–}{{\textendash}}1 {—}{{\textemdash}}1 {‘}{{\textquoteleft}}1 {’}{{\textquoteright}}1 {“}{{\textquotedblleft}}1 {”}{{\textquotedblright}}1]
<image><image>### Task Description
You are an expert in UI analysis. Please analyze the two given mobile GUI pages and
describe how a user operation causes the page to change from one state to another.
At the same time, identify and describe the UI elements required for the next
operation.

### Possible user operations and corresponding explanations
Possible user operations are represented as JSON dictionaries with an `action_type`
key. The other fields in the action dictionary are unique for each `action_type`
and  store parameter values for that action (e.g., "open_app" actions have an
"app_name" key, but no "x" or "y" keys). The `action_type` field will have one of
the following values:
* `click`: click at a specific point on the screen using the coordinates found in
the `'x'` and `'y'` fields of the action.
* `long_press`: long press at a specific point on the screen using the coordinates
found in the `'x'` and `'y'` fields of the action.
* `scroll`: scroll in a specific direction set in the `'direction'` field (one of
'up', 'down', 'left', or 'right').
* `open_app`: open an app with the name stored in the `'app_name'` field. This name
should map to an activity name which can be used to open the app using adb.
* `input_text`: types the text in the `'text'` field. It does not perform a click
or press enter at the end.
* `navigate_home`: go to the home screen.
* `navigate_back`: go back to the previous screen.
* `wait`: wait a set number of seconds. This is used when no action should be taken
other than to wait for something on screen (e.g., a loading bar).
Note: When it comes to position coordinates, please use a normalized coordinate
system between 0 and 1. Keep the coordinate values to four decimal places. The
coordinate origin is in the upper left corner of the image, with the X-axis
pointing to the right and the Y-axis pointing downward.

### Current user operation
The operation performed by the user on the current page is: {curr_action}

### Next user operation
The operation performed by the user on the next page is: {next_action}

### Task
Please describe in detail the page changes caused by this operation based on the
above information. If there are text elements on the GUI page, clearly indicate
what text may appear. If there are image elements on the GUI page, clearly describe
what images may appear. Please include the following:
1.Trigger Point of Change: Clearly indicate what operation of the user on the
current page triggered the page changes.
2.Elements on the Current Page: Describe in detail the UI elements of the current
page.
3.Elements on the Next Page: Describe in detail the UI elements of the next page.
4.Added Elements on the Page: Describe in detail which UI elements have been added
to the screen from Page 1 to Page 2. If there are no added UI elements, please
indicate so.
5.Reduced Elements on the Page: Describe in detail which UI elements have been
reduced on the screen from Page 1 to Page 2. If there are no reduced UI elements,
please indicate so.
6.Elements for Next Operation: Describe in detail the UI elements required for the
operation performed by the user on the next page and clearly indicate the position
of this UI element (e.g., {{\"index\":\"x,y\"}}). If the operation performed by the
user on the next page does not have the required UI elements, please still maintain
the structured output.

### Output format
Please output your analysis in a structured manner using the following title:
- ** Trigger operation **: [descriptions]
- ** Current elements **:
1. [descriptions 1]
2. [descriptions 2]
3. [descriptions 3]
continue...
- ** Next elements **:
1. [descriptions 1]
2. [descriptions 2]
3. [descriptions 3]
continue...
- ** Added elements **:
1. [descriptions 1]
2. [descriptions 2]
3. [descriptions 3]
continue...
- ** Reduced elements **:
1. [descriptions 1]
2. [descriptions 2]
3. [descriptions 3]
continue...
- ** Elements for next operation **:
{{\"coordinate\":\"x,y\",\"descriptions\":\"[elements descriptions]\"}}
\end{lstlisting}
\end{tcolorbox}

\subsection{Text-In Extraction for Trajectory Imagination}\label{sec:appendix-delta-text-prompt}

Given textual descriptions \texttt{\{ui\_elements\}} of the two consecutive pages (no images), this template identifies the trigger operation and enumerates the added, reduced, and next-operation elements, yielding the four core delta fields. It is applied to the UI-element descriptions designed by the App Designer (Appendix~\ref{sec:appendix-trajectories-prompt}) to obtain the delta-texts $\Delta_{1:T}$ that condition the world-model rollout during multi-step imagination, thereby constructing the agent's synthesized training data.

\begin{tcolorbox}[%
  title={Text-In Delta-Text Extraction Prompt},
  colback=white,
  colframe=black!80,
  coltitle=white,
  fonttitle=\bfseries\small,
  boxrule=0.4pt,
  arc=0pt,
  outer arc=0pt,
  left=2mm, right=2mm, top=1mm, bottom=1mm,
  width=\textwidth,
  breakable
]%
\begin{lstlisting}[breaklines=true,basicstyle=\footnotesize\ttfamily,columns=fullflexible,keepspaces=true,literate={–}{{\textendash}}1 {—}{{\textemdash}}1 {‘}{{\textquoteleft}}1 {’}{{\textquoteright}}1 {“}{{\textquotedblleft}}1 {”}{{\textquotedblright}}1]
You are an expert in UI analysis. Please analyze the UI elements descriptons on the
two given mobile GUI pages and describe how the first operation causes the GUI page
to change from one state to another. At the same time, identify and describe the UI
elements required for the second operation.

### Possible operations and corresponding explanations
Possible operations are represented as JSON dictionaries with an `action_type` key.
Other fields are **action-specific** and store parameter values required by that
action. Fields not relevant to a given `action_type` **must not appear** in the
action dictionary. The `action_type` field will have **one of the following
values**:

#### `status`
Reports the final status of the task.  
This action **ends the interaction**.
- **Fields**
  - `goal_status`: one of  
    - `"complete"` – the task has been successfully completed  
    - `"infeasible"` – the task cannot be completed

#### `answer`
Returns a textual response to the user (used when no further UI interaction is
needed).
- **Fields**
  - `text`: the answer content

#### `click`
Clicks at a specific point on the screen.
- **Fields**
  - `coordinate`: `x,y` normalized screen coordinates

#### `double_tap`
Performs a double tap at a specific point on the screen.
- **Fields**
  - `coordinate`: `x,y` normalized screen coordinates

#### `long_press`
Performs a long press at a specific point on the screen.
- **Fields**
  - `coordinate`: `x,y` normalized screen coordinates

#### `input_text`
Types the provided text at the given location.  
This action **does not** automatically press Enter.
- **Fields**
  - `text`: text to input
  - `coordinate`: `x,y` normalized screen coordinates

#### `keyboard_enter`
Simulates pressing the Enter / Return key on the keyboard.
- **Fields**
  - *(none)*

#### `scroll`
Scrolls the screen in a given direction.
- **Fields**
  - `direction`: one of `"up"`, `"down"`, `"left"`, `"right"`

#### `swipe`
Performs a swipe gesture starting from a specific point in a given direction.
- **Fields**
  - `coordinate`: `x,y` normalized screen coordinates
  - `direction`: one of `"up"`, `"down"`, `"left"`, `"right"`

#### `open_app`
Opens an application.
- **Fields**
  - `app_name`: name of the app to open (must map to an activity that can be
  launched via adb)

#### `navigate_home`
Navigates to the home screen.
- **Fields**
  - *(none)*

#### `navigate_back`
Navigates back to the previous screen.
- **Fields**
  - *(none)*

#### `wait`
Waits without performing any interaction.  
Used when the system needs time to update (e.g., loading, transitions).
- **Fields**
  - *(none)*

### Coordinate system note
- All screen positions use a **normalized coordinate system**:
  - Values range from **0 to 1**
  - Format: `x,y`
- Coordinates must be rounded to **four decimal places**
- Origin `(0, 0)` is the **upper-left corner** of the screen
- X-axis increases to the **right**
- Y-axis increases **downward**

### UI Elements Descriptons on the GUI Page
{ui_elements}

### UI Element Changes
Please describe in detail the UI element changes caused by the first operation
based on the above information. If there are text elements on the GUI page, clearly
indicate what text may appear. If there are image elements on the GUI page, clearly
describe what images may appear. Please complete the following task:
1. Trigger Point of Change: Clearly indicate what the first operation on the first
GUI page triggered the GUI page changes.
2. Added UI Elements: Describe in detail which UI elements have been added from GUI
Page 1 to GUI Page 2. If there are no added UI elements, please indicate so.
3. Reduced UI Elements: Describe in detail which UI elements have been reduced from
GUI Page 1 to GUI Page 2. If there are no reduced UI elements, please indicate so.
4. Elements for Second Operation: Describe in detail the UI elements required for
the operation on the GUI page 2 and clearly indicate the position of this UI
element (e.g., {{\"coordinate\":\"x,y\"}}). If the operation on the next page does
not have the required UI elements, please still maintain the structured output.

### Output format
Please output your answer in a structured manner using the following title:
- ** Trigger operation **: [descriptions]
- ** Added elements **:
1. [descriptions 1]
2. [descriptions 2]
3. [descriptions 3]
continue...
- ** Reduced elements **:
1. [descriptions 1]
2. [descriptions 2]
3. [descriptions 3]
continue...
- ** Elements for next operation **:
{{\"coordinate\":\"x,y\",\"descriptions\":\"[elements descriptions]\"}}
\end{lstlisting}
\end{tcolorbox}

\section{Trajectory Synthesis Templates}\label{sec:appendix-synthesis}

The two synthesis regimes of the Method section are each driven by a prompt template: single-frame appearance-only augmentation and multi-step imagination with an App Designer. Both are included below.

\subsection{Single-Frame Augmentation Prompt}\label{sec:appendix-augment-prompt}

Given a single screenshot, this template generates 3--5 appearance-only editing instructions that preserve the layout pixel-for-pixel. Each instruction draws from a distinct category---Color, Control Style, Icon, Background, or Theme---and alters only the visual skin (recoloring, icon or control restyling, background or theme replacement), leaving all text content and structure unchanged. Because only the visual skin is altered, the paired task instruction remains valid as-is.

\begin{tcolorbox}[%
  title={Single-Frame Augmentation Prompt},
  colback=white,
  colframe=black!80,
  coltitle=white,
  fonttitle=\bfseries\small,
  boxrule=0.4pt,
  arc=0pt,
  outer arc=0pt,
  left=2mm, right=2mm, top=1mm, bottom=1mm,
  width=\textwidth,
  breakable
]%
\begin{lstlisting}[breaklines=true,basicstyle=\footnotesize\ttfamily,columns=fullflexible,keepspaces=true,literate={–}{{\textendash}}1 {—}{{\textemdash}}1 {‘}{{\textquoteleft}}1 {’}{{\textquoteright}}1 {“}{{\textquotedblleft}}1 {”}{{\textquotedblright}}1]
You are a UI instruction generator for structure-consistent, appearance-only image
editing.

INPUT: A mobile GUI screenshot.
OUTPUT: Exactly 3 to 5 editing instructions. Each instruction must be a single,
self-contained line.

GLOBAL APPEARANCE-ONLY RULE:
All instructions must alter only the visual skin of the UI. The underlying
skeleton—layout, spacing, size, and position of every component—MUST remain
pixel-perfect to the original image.

TEXT CONTENT PRESERVATION RULE:
All displayed text (titles, labels, buttons, placeholders, etc.) must remain
exactly as shown. You may only change text color, font, or weight. The textual
string itself must never be altered.

ALLOWED VISUAL TRANSFORMATIONS:
- Recoloring any UI region or element.
- Replacing icons or images with new ones that share the same semantic purpose.
- Applying a new visual style to controls (e.g., changing a switch's appearance).
- Replacing the background layer.
- Applying an overall theme (e.g., Dark Mode).

FORBIDDEN STRUCTURAL CHANGES (do not generate instructions that imply):
- Adding, removing, moving, or resizing ANY element.
- Changing the text content.
- Altering margins, padding, or alignment.

ANALYSIS & DESCRIPTION RULES (for generating precise {placeholders}):
1. **Specificity:** When referring to a UI element, always use a distinguishing
feature (e.g., “the blue ‘Continue’ button at the bottom” not just “the button”).
2. **Style Hints for Icons:** When replacing an icon, include a brief style
descriptor to ensure visual coherence (e.g., “a flat, minimalist calendar icon” not
just “a calendar icon”).
3. **Color Descriptions:** Be specific. Instead of just “blue”, use a concrete
value or vivid description (e.g., “#2C3E50”, “a deep navy blue”, or “a vibrant
coral gradient”).

INSTRUCTION DIVERSITY RULE:
- Generate exactly 3, 4, or 5 instructions.
- Each instruction MUST use a DIFFERENT category from the list below.
- Prefer instructions that create a meaningful visual contrast (e.g., pairing a
color change with a style change) over superficial ones.

AVAILABLE CATEGORIES AND EXACT TEMPLATES (replace only {placeholders}):

[Color]
Change the color of {specific UI element with location/feature} to {specific color
value/description}.

[Control Style]
Change the style of {specific UI control with location/feature} to {target style
description}.

[Icon]
Replace {original icon description} in {specific location} with {new icon
description including style}.

[Background]
Replace the background with {new background description}.

[Theme]
Convert the UI to {target mode/theme}.

OUTPUT FORMAT RULES:
- One instruction per line. No markdown, no numbering, no bullet points.
- Separate instructions with exactly one blank line.
- Do NOT wrap the output in any way (no code blocks, no quotes).
- The output must contain only the instruction lines and the blank lines between
them.

EXAMPLE OUTPUT:
Change the color of the gradient header background to a solid deep indigo #1A1A2E.

Change the style of the prominent ‘Get Started’ button at the bottom to a
pill-shaped button with a soft shadow.

Replace the generic user avatar in the top-right profile card with a minimalist,
abstract geometric portrait icon.

Convert the UI to a high-contrast Dark Mode.

Now, based on the provided GUI image, generate the instructions.<image>\end{lstlisting}
\end{tcolorbox}

\subsection{Multi-Step Imagination (App Designer) Prompt}\label{sec:appendix-trajectories-prompt}

Given a task, a difficulty level, a target step count, and the initial page's UI elements, the App Designer imagines a coherent sequence of GUI pages and the operation on each. It outputs, per step, the designed UI elements, the step goal, and an operation drawn from the same action vocabulary as the text-in delta-text template, terminating with a \texttt{status: complete} action. The delta-texts conditioning the world-model rollout are then extracted from consecutive designed UI-element descriptions (Appendix~\ref{sec:appendix-delta-text-prompt}).

\begin{tcolorbox}[%
  title={Multi-Step Imagination (App Designer) Prompt},
  colback=white,
  colframe=black!80,
  coltitle=white,
  fonttitle=\bfseries\small,
  boxrule=0.4pt,
  arc=0pt,
  outer arc=0pt,
  left=2mm, right=2mm, top=1mm, bottom=1mm,
  width=\textwidth,
  breakable
]%
\begin{lstlisting}[breaklines=true,basicstyle=\footnotesize\ttfamily,columns=fullflexible,keepspaces=true,literate={–}{{\textendash}}1 {—}{{\textemdash}}1 {‘}{{\textquoteleft}}1 {’}{{\textquoteright}}1 {“}{{\textquotedblleft}}1 {”}{{\textquotedblright}}1]
You are a UI design expert. Please give full play to your imagination and
creativity. Given a task that can be completed on a mobile phone:<{task}>, with a
difficulty of <{difficulty}> and a number of steps of <{num_steps}>, design a
series of coherent, reasonable and rich GUI pages to simulate the process of
completing this task, and describe in detail the operations performed during the
process of completing this task.

### Possible operations and corresponding explanations
Possible operations are represented as JSON dictionaries with an `action_type` key.
Other fields are **action-specific** and store parameter values required by that
action. Fields not relevant to a given `action_type` **must not appear** in the
action dictionary. The `action_type` field will have **one of the following
values**:

#### `status`
Reports the final status of the task.  
This action **ends the interaction**.
- **Fields**
  - `goal_status`: one of  
    - `"complete"` – the task has been successfully completed  
    - `"infeasible"` – the task cannot be completed

#### `answer`
Returns a textual response to the user (used when no further UI interaction is
needed).
- **Fields**
  - `text`: the answer content

#### `click`
Clicks at a specific point on the screen.
- **Fields**
  - `coordinate`: `x,y` normalized screen coordinates

#### `double_tap`
Performs a double tap at a specific point on the screen.
- **Fields**
  - `coordinate`: `x,y` normalized screen coordinates

#### `long_press`
Performs a long press at a specific point on the screen.
- **Fields**
  - `coordinate`: `x,y` normalized screen coordinates

#### `input_text`
Types the provided text at the given location.  
This action **does not** automatically press Enter.
- **Fields**
  - `text`: text to input
  - `coordinate`: `x,y` normalized screen coordinates

#### `keyboard_enter`
Simulates pressing the Enter / Return key on the keyboard.
- **Fields**
  - *(none)*

#### `scroll`
Scrolls the screen in a given direction.
- **Fields**
  - `direction`: one of `"up"`, `"down"`, `"left"`, `"right"`

#### `swipe`
Performs a swipe gesture starting from a specific point in a given direction.
- **Fields**
  - `coordinate`: `x,y` normalized screen coordinates
  - `direction`: one of `"up"`, `"down"`, `"left"`, `"right"`

#### `open_app`
Opens an application.
- **Fields**
  - `app_name`: name of the app to open (must map to an activity that can be
  launched via adb)

#### `navigate_home`
Navigates to the home screen.
- **Fields**
  - *(none)*

#### `navigate_back`
Navigates back to the previous screen.
- **Fields**
  - *(none)*

#### `wait`
Waits without performing any interaction.  
Used when the system needs time to update (e.g., loading, transitions).
- **Fields**
  - *(none)*

### Coordinate system note
- All screen positions use a **normalized coordinate system**:
  - Values range from **0 to 1**
  - Format: `x,y`
- Coordinates must be rounded to **four decimal places**
- Origin `(0, 0)` is the **upper-left corner** of the screen
- X-axis increases to the **right**
- Y-axis increases **downward**

### Initial GUI page
## Step 0
{init_ui_elements}

### GUI page design and task planning
Based on the information above, You need to analyze the given task and plan the
operation steps for the user to complete this task. At the same time, exert your
imagination and creativity to design a series of coherent, reasonable and rich GUI
pages. Then, describe in detail the operations performed by the user during the
process of completing this task and the GUI pages you designed. If there are text
elements on the GUI page, clearly indicate what text may appear. If there are image
elements on the GUI page, clearly describe what images may appear. Please complete
the following task:
## Step id
1. UI Elements: Describe in detail all the UI elements on the GUI page you designed
from top to bottom and from left to right.
2. Step Goals: Describe in detail the goals you have planned for the current step
to complete the given task.
3. Operation: Clearly indicate the operations you have planned for the current step
to complete the given task (e.g., {{\"action_type\": \"click\", \"coordinate\":
\"x,y\"}}).

### Output format
Please output your answer in a structured manner using the following title:
## Step 0
{init_ui_elements}
- ** Step Goals **: [goals]
- ** Operation **: [operation] If the task is completed, then here is
{{\"action_type\": \"status\", \"goal_status\": \"complete\"}}.

## Step 1
- ** UI Elements **:
1. [descriptions 1]
2. [descriptions 2]
3. [descriptions 3]
continue...
- ** Step Goals **: [goals]
- ** Operation **: [operation] If the task is completed, then here is
{{\"action_type\": \"status\", \"goal_status\": \"complete\"}}.

## Step 2
- ** UI Elements **:
1. [descriptions 1]
2. [descriptions 2]
3. [descriptions 3]
continue...
- ** Step Goals **: [goals]
- ** Operation **: [operation] If the task is completed, then here is
{{\"action_type\": \"status\", \"goal_status\": \"complete\"}}.

continue...

## Step N
- ** UI Elements **:
1. [descriptions 1]
2. [descriptions 2]
3. [descriptions 3]
continue...
- ** Step Goals **: Task completed.
- ** Operation **: {{\"action_type\": \"status\", \"goal_status\": \"complete\"}}.\end{lstlisting}
\end{tcolorbox}

\section{Quality Filtering Details}\label{sec:appendix-relabel}

To retain only trajectories whose synthesized screenshots are actionable, we apply a VLM-based quality filter. For each generated step the VLM receives the task goal, the task progress, the intended step action, a reference action type, and the synthesized screenshot, and must emit a \texttt{mobile\_use} tool call of that same action type together with a confidence score from 0 to 100 reflecting how clearly the screenshot supports executing the action. Steps with low confidence---screenshots that lack the UI elements needed to carry out the reference action---are discarded, and trajectories containing such steps are removed from the training set. We set the threshold to $\eta = 90$ (on a 0--100 scale), discarding steps whose confidence falls below this value. The full filter template is included below.

\begin{tcolorbox}[%
  title={Quality Filtering Prompt},
  colback=white,
  colframe=black!80,
  coltitle=white,
  fonttitle=\bfseries\small,
  boxrule=0.4pt,
  arc=0pt,
  outer arc=0pt,
  left=2mm, right=2mm, top=1mm, bottom=1mm,
  width=\textwidth,
  breakable
]%
\begin{lstlisting}[breaklines=true,basicstyle=\footnotesize\ttfamily,columns=fullflexible,keepspaces=true,literate={–}{{\textendash}}1 {—}{{\textemdash}}1 {‘}{{\textquoteleft}}1 {’}{{\textquoteright}}1 {“}{{\textquotedblleft}}1 {”}{{\textquotedblright}}1]
You are an expert mobile UI analyst. Your task is to evaluate whether a generated
GUI screenshot contains sufficient and accurate information for an agent to execute
a specific action.

# ROLE & OBJECTIVE

Given:
1. A task objective (high-level goal)
2. Task progress (what has been completed so far)
3. A step action description (what the user should do next)
4. A reference action type (the TYPE of action that must be taken, e.g., click,
type, swipe, terminate)
5. A generated GUI screenshot

You must:
1. Analyze the screenshot to understand the current UI state
2. Determine the specific parameters for the reference action type based on the
screenshot
3. Output a tool_call with the SAME action type as the reference
4. Provide a confidence score (0-100) indicating how well the screenshot supports
the action

# OUTPUT FORMAT

Your response must include:

1. **Analysis**: Brief analysis of the screenshot and whether it supports the
reference action type
2. **Predicted Action**: A tool_call in XML format with the SAME action type as
provided in the reference
3. **Confidence**: A confidence score (0-100) based on:
   - How clear the UI elements are
   - Whether the screenshot matches the expected state for this action
   - How well the UI supports executing the action

# TOOL CALL FORMAT

The tool_call must be in this exact XML format:
<tool_call>
{"name": "mobile_use", "arguments": {"action": "ACTION_TYPE", ...other_fields...}}
</tool_call>

Where ACTION_TYPE can be: click, long_press, swipe, type, answer, system_button,
wait, terminate

For click/long_press: include "coordinate": [x, y]
For swipe: include "coordinate": [x1, y1], "coordinate2": [x2, y2]
For type: include "text": "..."
For system_button: include "button": "Back|Home|Menu|Enter"
For wait: include "time": seconds
For terminate: include "status": "success|failure"

# CONFIDENCE SCORING GUIDELINES

- **90-100**: Screenshot clearly shows all necessary UI elements, agent can
confidently execute the action
- **70-89**: Screenshot shows most necessary elements, minor ambiguity but action
is clear
- **50-69**: Screenshot shows some relevant elements but lacks clarity or has
missing information
- **30-49**: Screenshot is partially relevant but significant information is
missing
- **0-29**: Screenshot does not contain sufficient information or is irrelevant to
the task

# CRITICAL RULES

- You MUST output a tool_call with the SAME action type as the reference action
type provided
- DO NOT change the action type - if reference says "click", your output must be
"click"
- Determine the specific parameters (coordinates, text, etc.) based on the
screenshot content
- The confidence score reflects how well the screenshot supports the action
- Be realistic about what information is visible in the screenshot

# TASK OBJECTIVE
{task_goal}

# TASK PROGRESS
{task_progress}

# CURRENT STEP ACTION
{step_action}

# REFERENCE ACTION TYPE
{action_type}

# SCREENSHOT
<image>

---

Based on the screenshot above, analyze whether it contains sufficient information
for the agent to execute the described action. Provide:

1. **Analysis**: Is the screenshot appropriate for this step? Does it show the
necessary UI elements?
2. **Predicted Action**: Output a tool_call with action type "{action_type}" -
determine the specific parameters (coordinates, text, etc.) based on the
screenshot.
3. **Confidence**: How confident are you (0-100) that the predicted action is
correct?

Output format:
**Analysis**: [Your analysis here]

**Predicted Action**:
<tool_call>
{{"name": "mobile_use", "arguments": {{...}}}}
</tool_call>

**Confidence**: [0-100]\end{lstlisting}
\end{tcolorbox}

\subsection{Synthesis-Quality Evaluation Template}\label{sec:appendix-eval-template}

World-model synthesis quality (Table~\ref{tab:synthesis-quality-fidelity}) is judged by GLM-4.7v with the template below. Given a before screenshot, an after (generated) screenshot, a ground-truth after screenshot, and the editing instruction, the judge returns four 0--10 scores. We report these as Acc.\ (Accuracy), Comp.\ (Completeness), Cont.\ (Contextual alignment), and Vis.\ (Visual preservation) in the main table, matching its column headers.

\begin{tcolorbox}[%
  title={Synthesis-Quality Evaluation Prompt},
  colback=white,
  colframe=black!80,
  coltitle=white,
  fonttitle=\bfseries\small,
  boxrule=0.4pt,
  arc=0pt,
  outer arc=0pt,
  left=2mm, right=2mm, top=1mm, bottom=1mm,
  width=\textwidth,
  breakable
]%
\begin{lstlisting}[breaklines=true,basicstyle=\footnotesize\ttfamily,columns=fullflexible,keepspaces=true,literate={–}{{\textendash}}1 {—}{{\textemdash}}1 {‘}{{\textquoteleft}}1 {’}{{\textquoteright}}1 {“}{{\textquotedblleft}}1 {”}{{\textquotedblright}}1]
<image><image><image>You are an expert in evaluating the performance of a mobile
emulator. The mobile emulator is designed to navigate the UI change based on human
instruction. Given a series of instructions and the corresponding screenshots
before and after the action, your task is to assess how accurately the emulator has
followed the instructions.

### Input Format
- 'before' screenshot: The screenshot of the app's UI before the action is taken.
- 'after' screenshot: The screenshot of the app's UI after the action is taken.
- 'ground turth' screenshot: The turth screenshot of the app's UI after the action
is taken.
- instructions: A natural language description of the action that should be
performed on the app's UI before the action is taken.

### Input
- 'before' screenshot: The first screenshot.
- 'after' screenshot: The second screenshot.
- ''ground turth' screenshot': The third screenshot.
- instructions: {instructions_text}

### Task
Based on the information above, for each instruction, compare the 'before' and
'after' screenshots to determine if the action described in the instruction has
been correctly executed. Consider the following criteria in your evaluation:
1. Accuracy: Does the 'after' screenshot reflect the changes specified in the
instruction?
2. Completeness: Are all aspects of the instruction addressed in the 'after'
screenshot?
3. Contextual Understanding: Does the change make sense within the context of the
app's UI and functionality?
4. Visual Consistency: Are there any unintended changes in the 'after' screenshot
that were not part of the instruction?
Provide a detailed evaluation for each instruction, highlighting any discrepancies
or errors observed. Conclude with an overall assessment of the emulator's
performance in following the instructions. Please include the following:
** Reasons **: A reasoned analysis of the emulator's performance.
** Accuracy Metrics **: Specific metrics or scores that quantify the accuracy of
the emulator's actions. Score instructions on a scale from 0 to 10, where 10
indicates perfect execution.
** Completeness Metrics **: Specific metrics or scores that quantify how completely
the instructions were followed. Score instructions on a scale from 0 to 10, where
10 indicates full completeness.
** Contextual Understanding Metrics **: Specific metrics or scores that quantify
the emulator's understanding of the context. Score instructions on a scale from 0
to 10, where 10 indicates perfect contextual understanding.
** Visual Consistency Metrics **: Specific metrics or scores that quantify the
visual consistency of the 'after' screenshot. Score instructions on a scale from 0
to 10, where 10 indicates perfect visual consistency.

### Output Format
Please provide your evaluation in the following JSON format:
{{\"reasons\": <reasoned analysis>, \"metrics\": {{\"accuracy\": <accuracy score>,
\"completeness\": <completeness score>, \"contextual_understanding\": <contextual
understanding score>, \"visual_consistency\": <visual consistency score>}}}}
\end{lstlisting}
\end{tcolorbox}

\subsection{Error Propagation Dynamics in Recursive Rollouts}
\label{sec:appendix-error-propagation}

While Infinite-Dreamer achieves strong single-step fidelity (Table~\ref{tab:synthesis-quality-fidelity}), multi-step rollout quality exhibits a progressive decline over extended horizons (Table~\ref{tab:synthesis-quality-rollout}: from 70.71 at 1 step to 48.84 at 10 steps). Here we formalize the mathematical dynamics of error compounding during recursive generation, analyze the distribution shift between the two extraction templates, and describe the protective role of VLM-based quality filtering.

\paragraph{Formal Model of Compounding Drift.}
Let $s_t^*$ denote the ground-truth reachable GUI screenshot at step $t$, and let $\tilde{s}_t = s_t^* + \mathbf{e}_t^{\mathrm{vis}}$ denote the generated observation produced by recursive application of $f_\phi$. Similarly, let $\Delta_t^*$ denote the ideal transition specification, and let $\tilde{\Delta}_t = \Delta_t^* + \mathbf{e}_t^{\mathrm{sem}}$ denote the delta-text authored by the App Designer. The visual state at step $t+1$ evolves as:
\begin{equation}
\tilde{s}_{t+1} = f_\phi(\tilde{s}_t, \tilde{\Delta}_t) = f_\phi(s_t^* + \mathbf{e}_t^{\mathrm{vis}},\, \Delta_t^* + \mathbf{e}_t^{\mathrm{sem}}).
\end{equation}
Taking a first-order Taylor expansion around the ground-truth transition $(s_t^*, \Delta_t^*)$, the visual error vector $\mathbf{e}_{t+1}^{\mathrm{vis}}$ at step $t+1$ satisfies:
\begin{equation}
\mathbf{e}_{t+1}^{\mathrm{vis}} = \tilde{s}_{t+1} - s_{t+1}^* \approx \underbrace{\nabla_s f_\phi(s_t^*, \Delta_t^*) \cdot \mathbf{e}_t^{\mathrm{vis}}}_{\text{Autoregressive Error Compounding}} + \underbrace{\nabla_\Delta f_\phi(s_t^*, \Delta_t^*) \cdot \mathbf{e}_t^{\mathrm{sem}}}_{\text{Semantic Mismatch Distortion}} + \mathbf{\delta}_{\mathrm{flow}},
\label{eq:error-propagation}
\end{equation}
where $\nabla_s f_\phi$ is the Jacobian of the world model with respect to the input image, $\nabla_\Delta f_\phi$ captures sensitivity to delta-text specifications, and $\mathbf{\delta}_{\mathrm{flow}}$ represents numerical integration error from flow-matching sampling.

Equation~\ref{eq:error-propagation} reveals two distinct driving sources of rollout error:
\begin{enumerate}
    \item \textbf{Semantic Specification Shift ($\mathbf{e}_t^{\mathrm{sem}}$):} During world-model training, the extractor $E_\psi$ consumes ground-truth image pairs $(s_t, s_{t+1})$ to extract $\Delta_t$. During multi-step imagination, however, the text-only extractor $E_\psi^{\mathrm{txt}}$ must infer $\Delta_t$ solely from high-level UI descriptions authored by the App Designer $G_\omega$, without ground-truth visual feedback. When $E_\psi^{\mathrm{txt}}$ underspecifies secondary UI effects (e.g., background dimming or keyboard dismissal), $\mathbf{e}_t^{\mathrm{sem}}$ induces visual distortions in the generated frame.
    \item \textbf{Visual Memory Amplification ($\mathbf{e}_t^{\mathrm{vis}}$):} Once $\tilde{s}_t$ departs from the manifold of clean training screenshots, the corrupted canvas is fed back into $f_\phi$ at step $t+1$. Because diffusion models lack explicit self-correcting mechanisms, slight font blurring or boundary shifts are often misinterpreted by $f_\phi$ as intentional UI structures and further amplified by $\nabla_s f_\phi$.
\end{enumerate}

\paragraph{Rollout Survival under Quality Filtering.}
To characterize how error compounding affects synthesized data, we measure two quantities on the validation cache of the multi-step imagination pipeline ($1{,}599$ imagined trajectories, $9{,}504$ validated generated screenshots; the initial page at step~0 is excluded). At horizon $t$ we report the fraction of evaluated steps at that horizon that pass the VLM quality filter ($\eta \ge 90$), and the cumulative fraction of trajectories reaching horizon $t$ whose steps $1,\ldots,t$ all pass. The at-risk set falls from $1{,}599$ trajectories at $t=1$ to $302$ at $t=10$, because imagined tasks end once their final action is reached ($91\%$ of trajectories end with a \texttt{terminate} or \texttt{answer} action).

\begin{table}[h]
\centering
\caption{Survival of synthesized rollouts under VLM quality filtering ($\eta = 90$) across horizons $t=1,\dots,10$, measured over $1{,}599$ imagined trajectories. \emph{Step validity}: share of evaluated steps at horizon $t$ that pass. \emph{Cumulative survival}: share of trajectories reaching horizon $t$ whose steps $1,\ldots,t$ all pass. Step~0 (the initial page) is excluded.}
\label{tab:rollout-survival}
\small
\setlength{\tabcolsep}{4.5pt}
\renewcommand{\arraystretch}{1.15}
\begin{tabular}{lcccccccccc}
\toprule
\textbf{Rollout Step ($t$)} & \textbf{1} & \textbf{2} & \textbf{3} & \textbf{4} & \textbf{5} & \textbf{6} & \textbf{7} & \textbf{8} & \textbf{9} & \textbf{10} \\
\midrule
Step Validity (\%) & 84.6 & 84.9 & 82.8 & 84.7 & 79.0 & 78.4 & 80.4 & 79.4 & 81.2 & 79.5 \\
Cumulative Survival (\%) & 84.6 & 72.0 & 60.6 & 50.3 & 39.8 & 30.4 & 24.1 & 18.7 & 14.1 & 9.3 \\
\bottomrule
\end{tabular}
\end{table}

Step validity stays within a narrow band ($78.4$--$84.9\%$) across all horizons, so the filter is not systematically stricter on later steps. Cumulative survival, by contrast, falls from $84.6\%$ at step~1 to $9.3\%$ at step~10, because a trajectory remains usable only while every one of its steps passes. The VLM quality filter therefore acts as a \emph{circuit-breaker}: compounding degradation, rather than a collapsing per-step pass rate, is what removes trajectories, and filtering keeps those with degraded UI elements out of the final policy training set $\mathcal{D}_{\mathrm{gen}}$.

\section{Additional Qualitative Examples}\label{sec:appendix-qualitative}

This section provides qualitative comparisons between Infinite-Dreamer and baseline image-editing models across single-step and multi-step scenarios.

\subsection{Single-Step Editing Comparison}
\label{sec:appendix-single-step-comparison}

\begin{figure}[t]
\centering
\includegraphics[width=\textwidth]{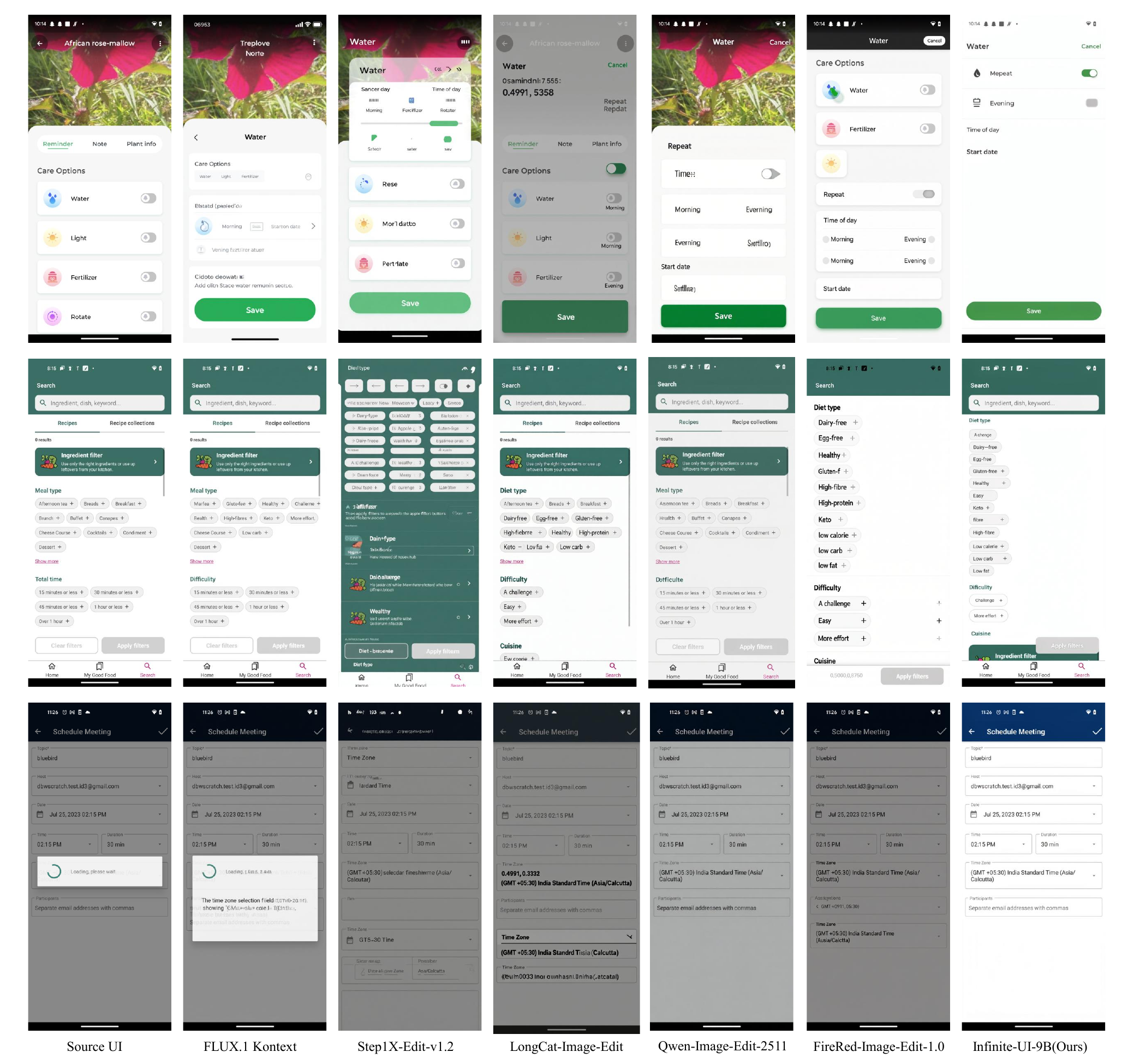}
\caption{Qualitative comparison of single-step UI editing. Three examples (rows) show the source UI and outputs from FLUX.1 Kontext, Step1X-Edit-v1.2, LongCat-Image-Edit, Qwen-Image-Edit-2511, FireRed-Image-Edit-1.0, and Infinite-Dreamer-9B (rightmost). Infinite-Dreamer-9B most faithfully executes the requested edits while preserving layout and unmodified regions.}
\label{fig:appendix-single-step-G1}
\end{figure}

We present three representative single-step editing examples comparing Infinite-Dreamer-9B against several baseline image-editing models (FLUX.1 Kontext, Step1X-Edit-v1.2, LongCat-Image-Edit, Qwen-Image-Edit-2511, FireRed-Image-Edit-1.0). The examples cover a plant-care settings screen, a recipe-filter page, and a schedule-meeting form. For each example, the source UI is transformed according to a delta-text instruction specifying UI elements to add, remove, or modify. As shown in Figure~\ref{fig:appendix-single-step-G1}, Infinite-Dreamer-9B produces edits that more accurately reflect the intended transitions while preserving the overall UI structure and unmodified regions. Baseline models often fail to apply the requested changes correctly or introduce artifacts in unedited regions.

\subsection{Multi-Step Rollout Comparison}
\label{sec:appendix-multi-step-comparison}

Figure~\ref{fig:appendix-multi-step-G2} shows a five-step rollout in the LeafSnap plant app across all compared models. Part~I presents three general-purpose baselines (FLUX.1 Kontext, Step1X-Edit-v1.2, LongCat-Image-Edit), Part~II compares Qwen-Image-Edit-2511, FireRed-Image-Edit-1.0, and Infinite-Dreamer-4B (Ours), and Part~III shows the ground-truth trajectory alongside Infinite-Dreamer-9B (Ours). Our models faithfully track the ground-truth UI state transitions, while baselines accumulate compounding errors across steps.

\begin{figure}[t]
\centering
\includegraphics[width=\textwidth]{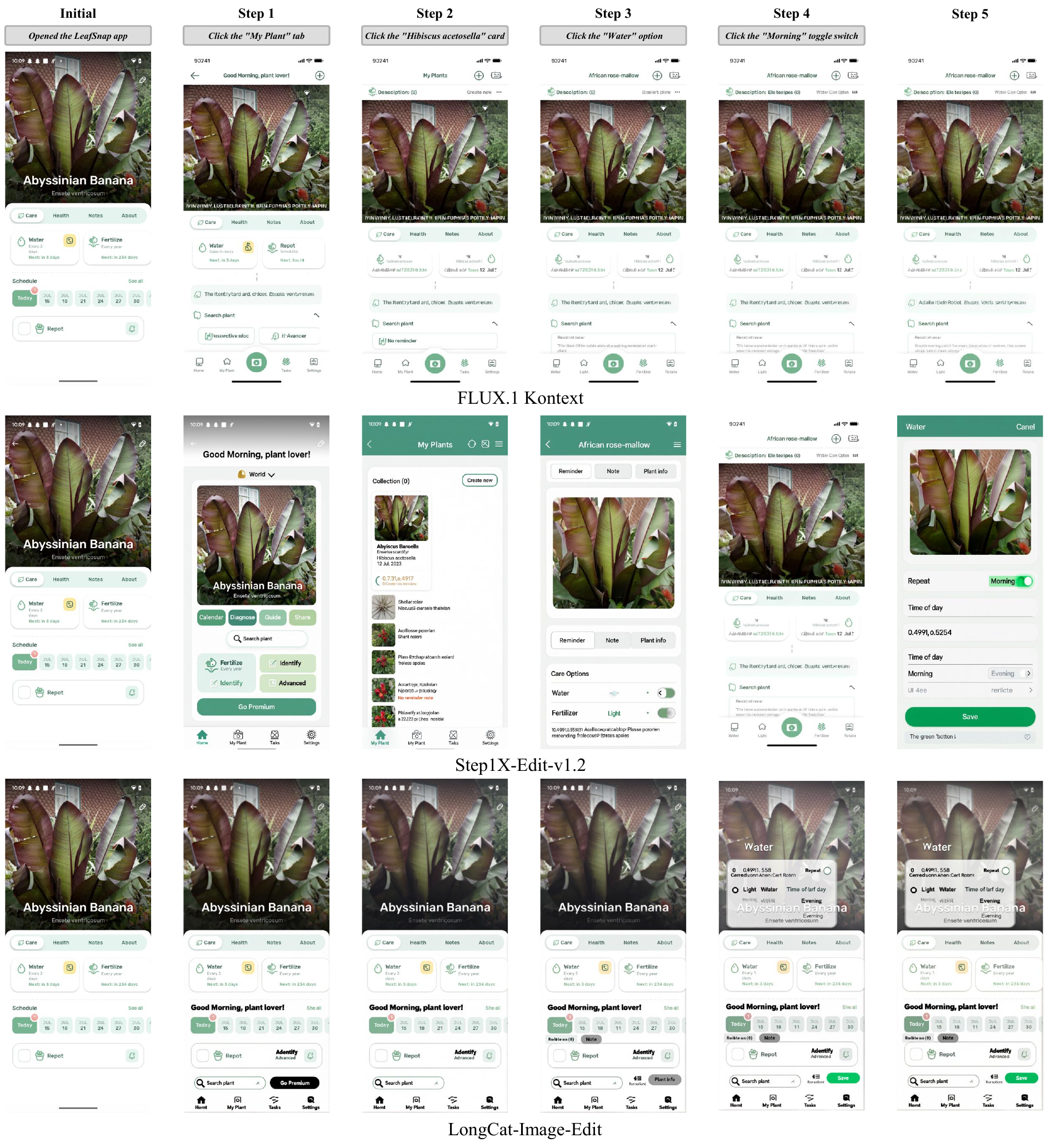}
\caption{Qualitative comparison of multi-step rollout. (a)~Part~I shows outputs from FLUX.1 Kontext, Step1X-Edit-v1.2, and LongCat-Image-Edit for the five-step LeafSnap watering task; each column is a step. These baselines drift away from the intended UI state.}
\label{fig:appendix-multi-step-G2}
\end{figure}

\begin{figure}[t]
\centering
\includegraphics[width=\textwidth]{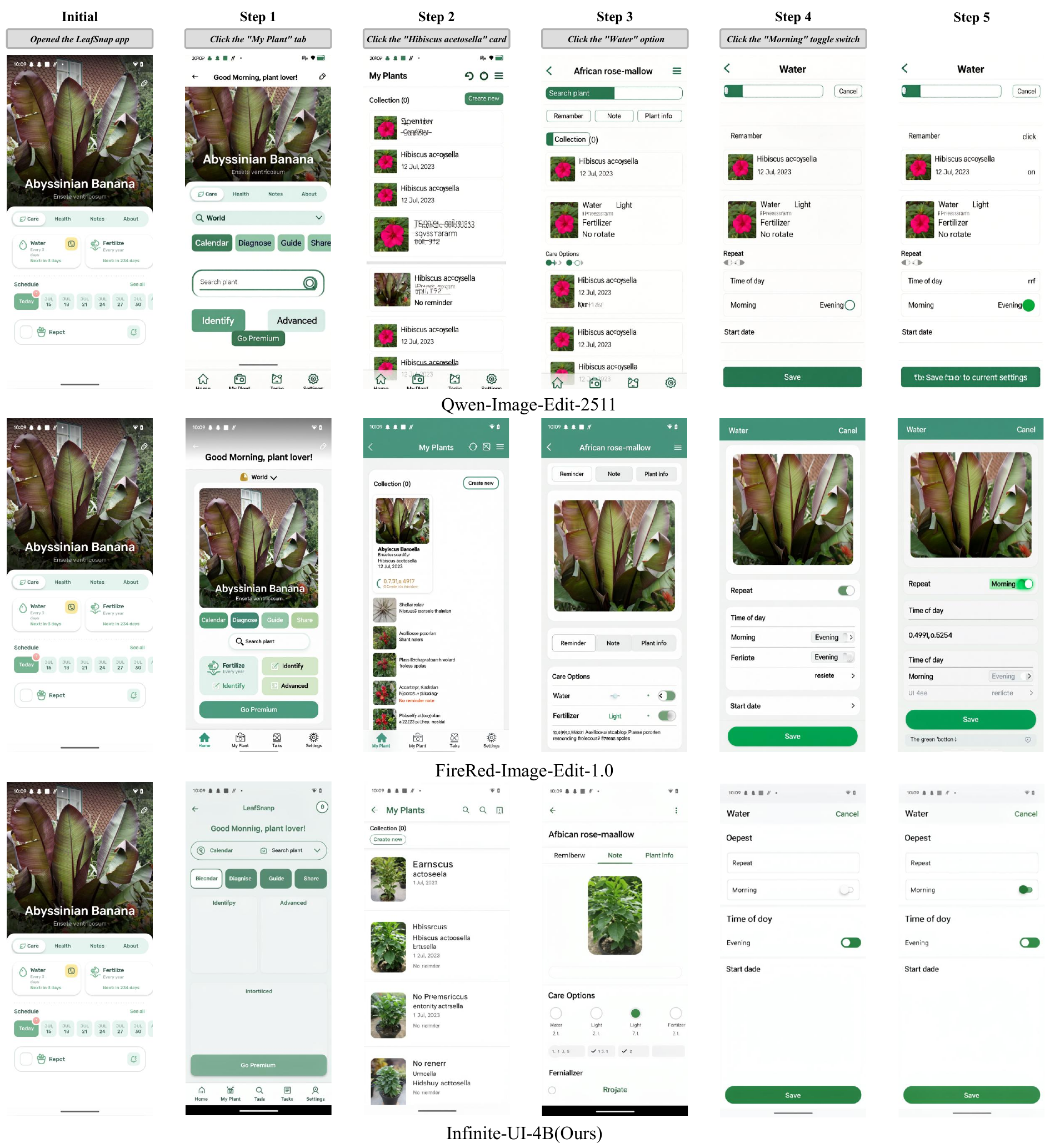}
\addtocounter{figure}{-1}
\caption{(b)~Part~II shows outputs from Qwen-Image-Edit-2511, FireRed-Image-Edit-1.0, and Infinite-Dreamer-4B (Ours) for the same rollout. Infinite-Dreamer-4B preserves visual coherence while the baselines introduce artifacts and layout drift.}
\end{figure}

Across all five steps, our Infinite-Dreamer models preserve plant names, UI text, and the toggled reminder state correctly, while baselines show corrupted text, duplicated or missing UI elements, and inconsistent backgrounds---demonstrating the importance of domain-specific fine-tuning for GUI state synthesis.

\begin{figure}[t]
\centering
\includegraphics[width=\textwidth]{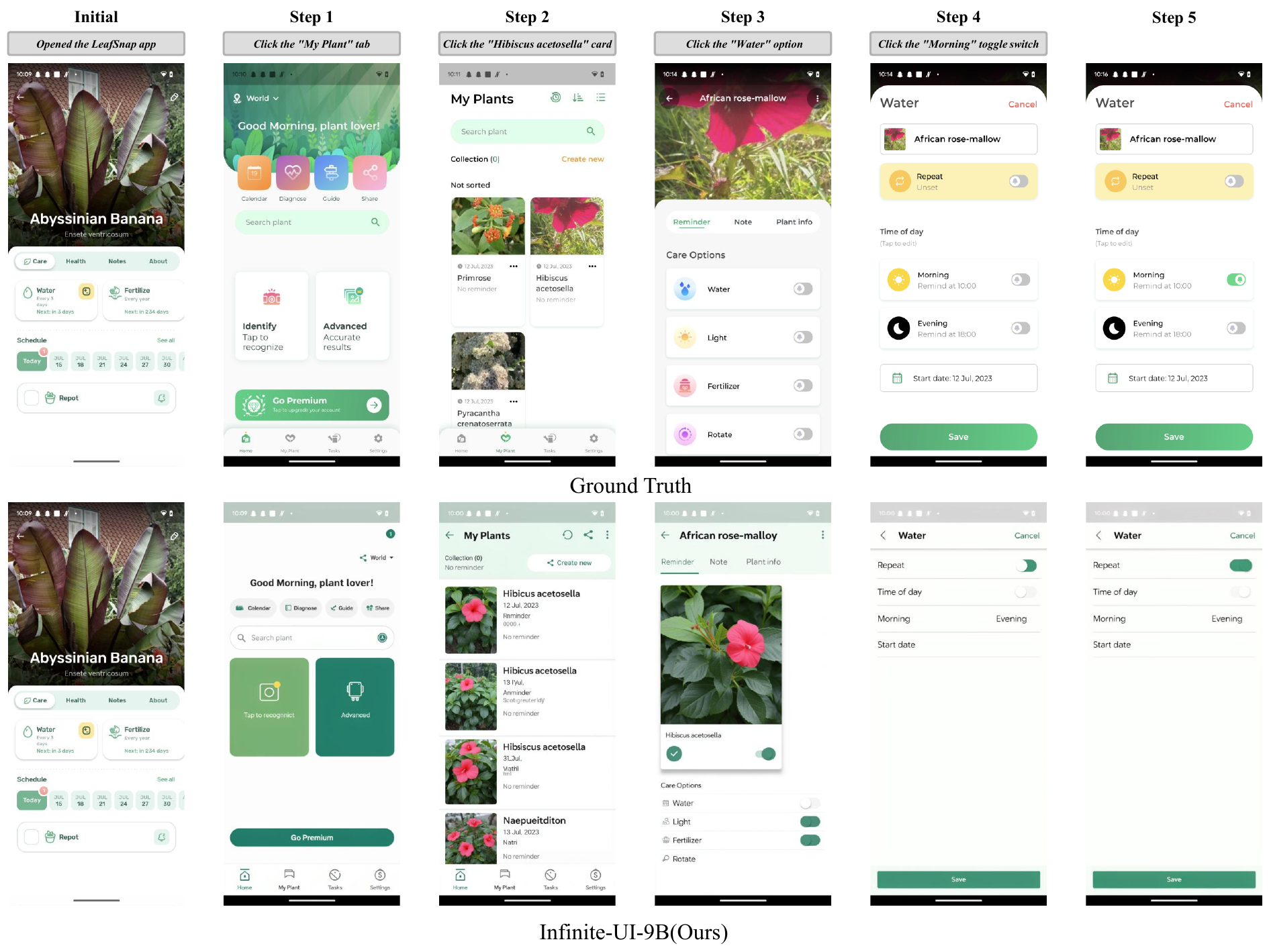}
\addtocounter{figure}{-1}%
\caption{(c)~Part~III compares the ground-truth trajectory with Infinite-Dreamer-9B (Ours). Infinite-Dreamer-9B closely follows the ground-truth layout and text across all five steps.}
\end{figure}

\section{More Visualizations}
\label{sec:appendix-more-visualizations}

This section showcases the diversity and quality of images generated by Infinite-Dreamer under two synthesis regimes: single-frame visual augmentation and multi-step trajectory imagination.

\subsection{Single-Frame Augmentation Results}
\label{sec:appendix-augmentation-results}

\begin{figure}[p]
\centering
\includegraphics[width=\textwidth]{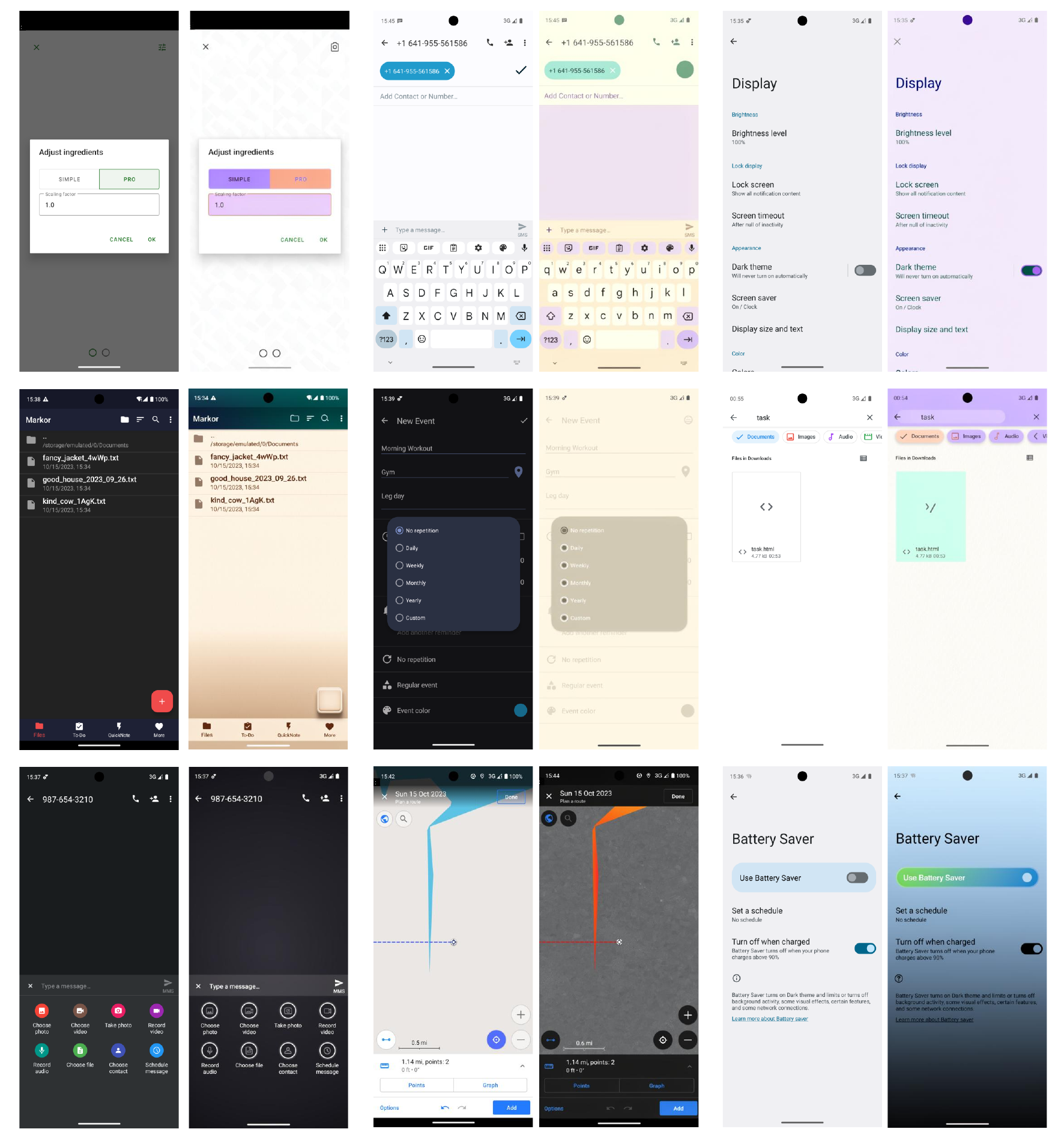}
\caption{Single-frame augmentation examples. For each pair, the left column shows the source screenshot and the right column shows a visually distinct variant produced by Infinite-Dreamer. The pairs span multiple app categories and cover dark/light theme swap, accent-color and background recoloring, and icon/keyboard restyling, while widget positions and text content are held fixed.}
\label{fig:augmentation-results}
\end{figure}

Single-frame augmentation generates visually diverse variants of a source screenshot while strictly preserving the layout and all text content. Figure~\ref{fig:augmentation-results} shows source--variant pairs spanning multiple app categories, including a recipe editor, a dialer, a system Display settings page, a file manager, an event scheduler, a file-explorer document preview, a messaging composer, a hiking route planner, and a Battery Saver settings page. The variants exercise dark/light theme swap, accent-color and background recoloring, and icon/keyboard restyling, while every button, list item, and text label from the source is reproduced verbatim. These augmentations improve the agent's robustness to surface-level UI variations without requiring new task labels.

\subsection{Multi-Step Imagination Results}
\label{sec:appendix-imagination-results}

\begin{figure}[p]
\centering
\includegraphics[width=\textwidth]{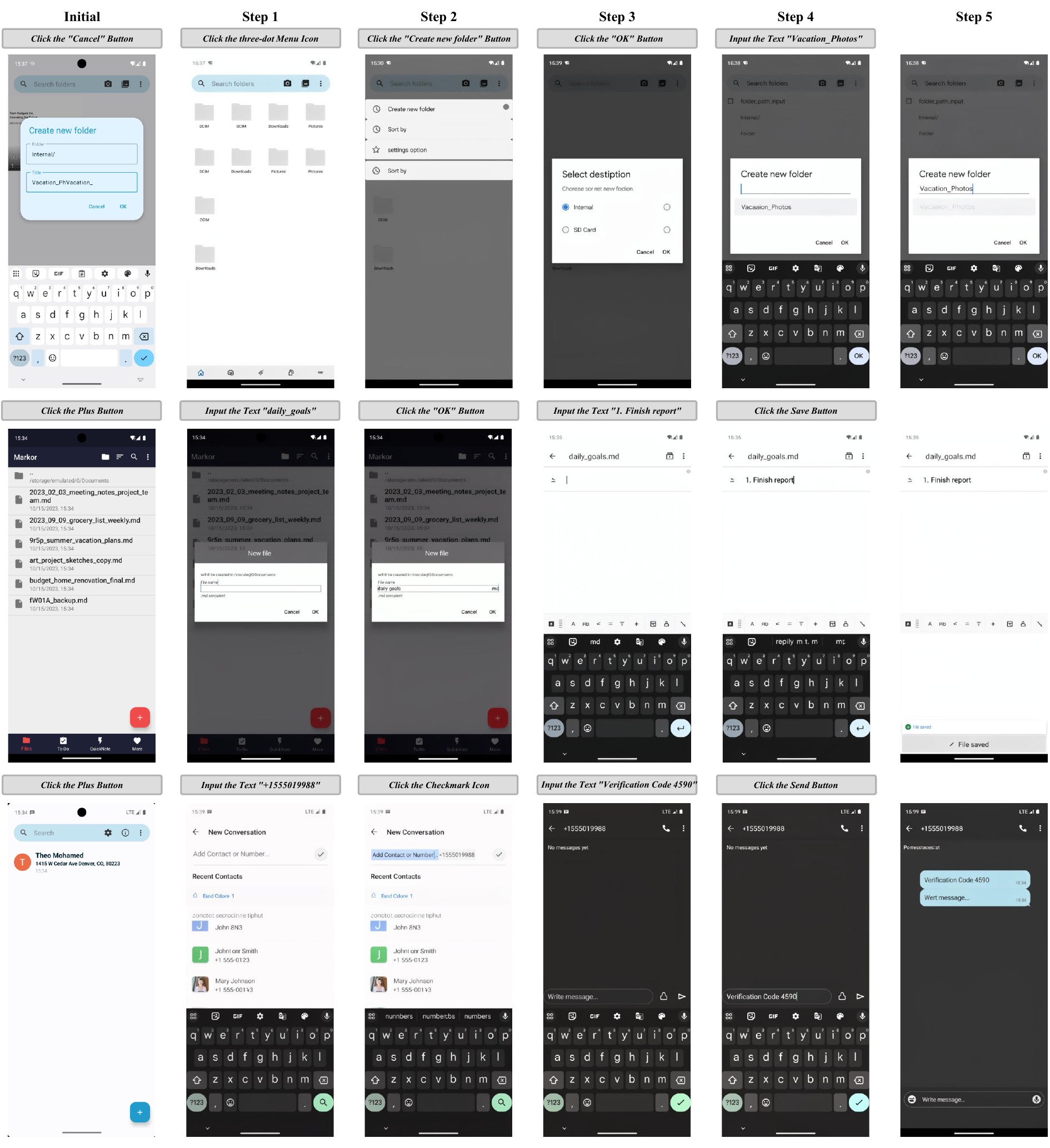}
\caption{Multi-step imagination examples. Each row is an independent trajectory: starting from an Initial screenshot, Infinite-Dreamer autoregressively applies the delta-text action printed above each column (e.g., ``Click the \emph{Create new folder} Button'', ``Input the Text \emph{daily\_goals}'', ``Click the Send Button'') to produce Step~1 through Step~5. The imagined screenshots keep the surrounding UI chrome stable while correctly rendering the widget or text that the action introduces.}
\label{fig:imagination-results}
\end{figure}

Multi-step imagination generates complete trajectory rollouts from a seed screenshot using the App Designer and world model. Figure~\ref{fig:imagination-results} shows three independent rollouts, each starting from an Initial state and unfolding over five steps. Every step is conditioned on the previous screenshot together with a delta-text action label---for example ``Click the \emph{Create new folder} Button'', ``Input the Text \emph{daily\_goals}'', or ``Click the Send Button''. Across the three trajectories, Infinite-Dreamer opens dialogs, types into text fields, and dismisses modals while preserving the surrounding UI chrome.

\subsection{Detailed Failure Mode Analysis, Diagnostics, and Mitigations}
\label{sec:appendix-failure-cases}

While Infinite-Actor demonstrates marked improvements on end-to-end task success (AndroidWorld Pass@1 +4.45 to +9.05), two performance anomalies warrant deeper investigation: (i) regressions in coordinate grounding (Box Acc) on AC-Curated, and (ii) a -2.7 point regression on AndroidWorld Hard tasks when mixing synthetic data with real demonstrations (Figure~\ref{fig:difficulty-comparison}). Here we measure the failure outcomes of the mixed-data model from its evaluation logs, relate them to tier-dependent regressions, and discuss candidate mechanisms together with potential mitigations.

\paragraph{Measured Failure Outcomes on AndroidWorld Hard Tasks.}
To understand why synthetic data fails to boost Hard tasks, we analyze the evaluation logs of \texttt{mix-4b} on the 19 AndroidWorld Hard tasks (one episode per task: two successes and 17 failures). Each failure is assigned to exactly one terminal outcome recorded in the log and listed in Table~\ref{tab:hard-task-errors}. The episode either exhausted the interaction budget the harness allocates ($10\times$ the task complexity score), terminated with \texttt{goal\_status}=\texttt{complete} although the verifier still failed the task, ended with an \texttt{answer} action that failed verification, or stopped mid-interaction without terminating. None of the 17 failures ends with an explicit infeasible verdict.

\begin{table}[h]
\centering
\caption{Measured failure outcomes of \texttt{mix-4b} on the 19 AndroidWorld Hard tasks (17 failures, one episode per task). Categories are mutually exclusive and are assigned from the terminal event recorded in the evaluation log.}
\label{tab:hard-task-errors}
\small
\setlength{\tabcolsep}{5pt}
\renewcommand{\arraystretch}{1.15}
\begin{tabular}{p{3.6cm}cp{7.4cm}}
\toprule
\textbf{Failure Mode} & \textbf{Share} & \textbf{Operational Definition (from the evaluation log)} \\
\midrule
Step-budget exhaustion & 41.2\% (7/17) & Ran to the allocated interaction budget ($10\times$ task complexity) while still acting; no terminal action was issued. \\
\midrule
False completion & 35.3\% (6/17) & Issued a \texttt{terminate} call with \texttt{goal\_status}=\texttt{complete}, yet the task verifier marked the episode failed. \\
\midrule
Incorrect final answer & 17.6\% (3/17) & Ended with an \texttt{answer} action whose response failed verification (three information-retrieval tasks). \\
\midrule
Abrupt stop & 5.9\% (1/17) & Episode ended mid-interaction; the log records no \texttt{terminate} or \texttt{answer} call. \\
\bottomrule
\end{tabular}
\end{table}

Two process-level symptoms accompany these outcomes. First, $6/17$ failed episodes contain a run of at least three consecutive identical actions, meaning the agent repeats an action or oscillates between screens instead of progressing, and the same number of episodes contain the agent's own statement that it is stuck in a loop. Second, $7/17$ episodes execute at least one action after which the screenshot is byte-identical, and $3/17$ issue clicks that produce no screen change at all. These symptoms concentrate in the Hard tier: among failed episodes, false completion rises from $14.3\%$ (Easy) and $30.4\%$ (Medium) to $35.3\%$ (Hard), and repetition or oscillation rises from $19.0\%$ and $21.7\%$ to $35.3\%$. Both Hard successes do terminate with \texttt{goal\_status}=\texttt{complete} and pass verification, so the discriminating factor is not whether the agent terminates but whether its claimed state matches the environment.

\begin{figure}[h]
\centering
\includegraphics[width=\textwidth]{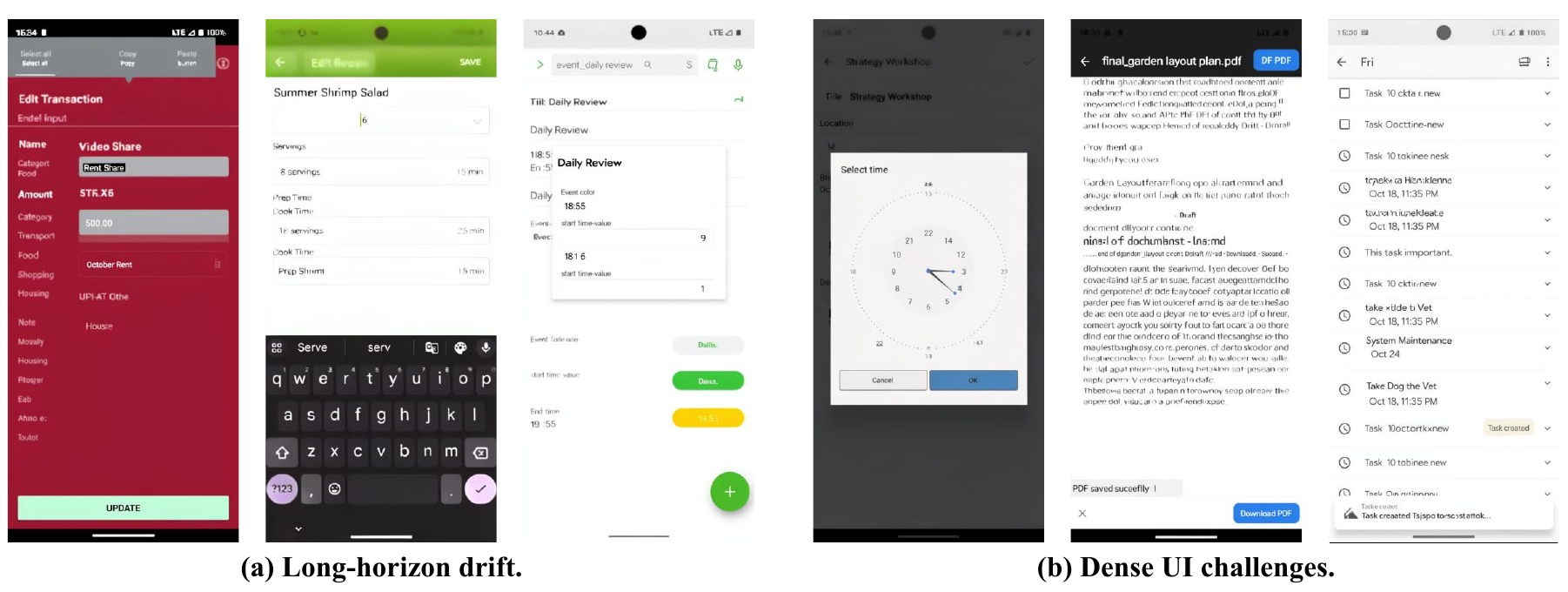}
\caption{Representative failure cases of Infinite-Dreamer. (a)~Long-horizon drift: over many generated steps, visual inconsistencies such as garbled text, shifted buttons, and spurious elements accumulate. (b)~Dense UI challenges: tightly packed widgets (e.g., grid layouts, complex lists) can cause the model to merge adjacent elements or misplace insertions and removals.}
\label{fig:appendix-failure-cases}
\end{figure}

\paragraph{From Measured Outcomes to Mechanisms.}
Table~\ref{tab:hard-task-errors} describes how Hard episodes fail, not which property of the synthetic corpus caused the failure. We consider four candidate mechanisms; each would require a controlled study to confirm. First, \emph{coordinate displacement}: on AC-Curated, Infinite-Actor-4B's predicted regions are displaced from the annotations rather than marginally noisy, and the loss concentrates on compact targets (Section~\ref{sec:experiments}). That comparison changes both fine-tuning and the data sources, so it is consistent with, but does not isolate, an effect of synthetic supervision; separating boundary changes in the generated screenshots from the policy's visual encoding would require a controlled study. Its behavioural symptom in the Hard logs would be ineffective clicks, which occur in $3/17$ failures. Second, \emph{long-horizon causal noise}: world-model fidelity decays with rollout length (Table~\ref{tab:synthesis-quality-rollout}), and only $9.3\%$ of trajectories that reach step~10 survive quality filtering intact (Table~\ref{tab:rollout-survival}), so long synthetic rollouts are both scarce and biased toward easy horizons. A policy trained on such supervision can adopt over-optimistic beliefs about progress and termination, which is consistent with the dominance of budget exhaustion and false completion above. Third, \emph{dense-UI confusion}: as shown in Figure~\ref{fig:appendix-failure-cases}b, the world model can merge adjacent widgets or misplace insertions in tightly packed layouts, which in principle teaches incorrect element attributes. Fourth, \emph{transient system-dialog inconsistency}: ephemeral runtime permission dialogs and system keyboards are unlikely to appear in imagined screenshots, so a policy trained on such data may mishandle them on a real device. The evaluation logs record actions but not whether a dialog was on screen, and they contain no element-level annotations, so the third and fourth mechanisms are not measurable from them and remain hypotheses.

\paragraph{Constructive Mitigation Strategies.}
These findings point to actionable avenues for advancing generative GUI world models:
\begin{itemize}
    \item \textbf{Termination calibration and progress verification:} Since false completion accounts for $35.3\%$ of Hard failures, training should verify task post-conditions before accepting a \texttt{terminate} call, for example by re-checking the environment state or by filtering synthetic trajectories whose terminal state was never observed in real data.
    \item \textbf{Box-preserving attention guidance:} Incorporating explicit bounding-box coordinate losses or masking during world-model fine-tuning can anchor element boundaries more tightly and reduce the coordinate displacement of predicted regions.
    \item \textbf{Keyframe re-anchoring in long horizons:} Since only $9.3\%$ of rollouts that reach step~10 remain fully valid, periodically re-anchoring the visual state to real seed screenshots or a template library can prevent compounding visual and causal drift before it exhausts the horizon.
    \item \textbf{Consistency verification for Hard tasks:} Applying bi-directional rollout verification (e.g., checking whether reverse actions reconstruct the prior state) before admitting long-horizon synthetic trajectories into the training corpus can suppress spurious transitions.
\end{itemize}

\FloatBarrier

\end{document}